%% file: Bayes_v2025.tex
\documentclass[11pt]{article}
\usepackage{fullpage}

\usepackage[utf8]{inputenc} 
\pdfoutput=1

\usepackage{fullpage}
\usepackage{natbib}
\usepackage{graphicx}
\usepackage{subcaption}
\usepackage{import}

\usepackage{amsmath, amssymb, amsfonts, mathtools, bm, bbm, mathrsfs}
\usepackage{algorithm, algorithmic}
\usepackage{booktabs, multirow, makecell, threeparttable}
\usepackage{enumerate, enumitem}
\usepackage{xcolor, color}
\usepackage{verbatim, comment}
\usepackage{array}
\usepackage{appendix}
\usepackage{footnote}
\usepackage{ragged2e}
\usepackage{microtype}
\usepackage{hyperref}
\usepackage{dsfont}

\hypersetup{
	colorlinks=true,
	citecolor = blue,
	filecolor=blue,
	urlcolor=cyan,
}

\newtheorem{theorem}{Theorem}
\newtheorem{corollary}{Corollary}
\newtheorem{lemma}{Lemma}

\newtheorem{definition}{Definition}

\newtheorem{remark}{Remark}
\newtheorem{proof}{Proof}

\DeclareMathOperator*{\argmax}{argmax}
\DeclareMathOperator*{\argmin}{argmin}

\def\bs{\boldsymbol}
\def\nohao{\nonumber}

\title{Bayes-Optimal Fair Classification with Multiple Sensitive Features}

\date{November, 2025}

\author{
	Yi Yang\thanks{Department of Information Systems, Arizona State University; E-mail:  \texttt{Yi.Yang.10@asu.edu}} \\
	\and
	Yinghui Huang\thanks{Corresponding Author.} \thanks{Department of Information Systems and Intelligent Business,	
    School of Management, Xi'an Jiaotong University; 
    E-mail: \texttt{yinghui.huang@xjtu.edu.cn}. }  \\
	\and
	Xiangyu Chang
    \thanks{Department of Information Systems and Intelligent Business, School of Management, Xi’an Jiaotong University; E-mail: \texttt{xiangyuchang@xjtu.edu.cn}. }
}

\begin{document}

\maketitle

\begin{abstract}
Existing theoretical work on Bayes-optimal fair classifiers usually considers a single (binary) sensitive feature.
In practice, individuals are often defined by multiple sensitive features. 
In this paper, we characterize 
the Bayes-optimal fair classifier for multiple sensitive features under general approximate fairness measures, including \textit{mean difference} and \textit{mean ratio}.
We show that these approximate measures for existing group fairness notions, including Demographic Parity, Equal Opportunity, Predictive Equality, and Accuracy Parity, are linear transformations of selection rates for specific groups defined by both labels and sensitive features.
We then characterize that Bayes-optimal fair classifiers for multiple sensitive features become instance-dependent thresholding rules that rely on a weighted sum of these group membership probabilities.
Our framework applies to both attribute-aware and attribute-blind settings and can accommodate composite fairness notions like Equalized Odds. 
Building on this, we propose two practical algorithms for Bayes-optimal fair classification via in-processing and post-processing. 
We show empirically that our methods compare favorably to existing methods.
\end{abstract}

\footnotetext[1]{This paper has been accepted to the AAAI-26 main track.}
\import{./}{1-Introduction.tex}

\import{./}{2-Preliminaries.tex}

\import{./}{3-Genernal_fair_measure.tex}

\import{./}{4-Bayes-optimal_fair_classifier.tex}

\import{./}{5-Algorithm.tex}

\import{./}{6-Experiments.tex}

\import{./}{7-Conclusion.tex}

\bibliographystyle{apalike}
\bibliography{example_paper} 

\newpage
\section*{Appendix}

\begin{appendix} \label{appendix}
\import{./}{8-Appendix.tex}	
\end{appendix}

\end{document}

%% file: 1-Introduction.tex
\section{Introduction}\label{sec:Introduction}
Machine learning (ML) models have become integral to decision-making processes in various high-stakes fields, such as credit scoring and criminal justice. 
However, a growing concern has emerged regarding the fairness of these models, particularly with respect to outputs that may disadvantage certain social groups defined by sensitive features such as race, gender, or socio-economic status \citep{barocas2023}. 
Therefore, addressing fairness issues in ML has garnered significant attention \citep{mehrabi2021survey,caton2024fairness}.

A considerable body of work has focused on fairness in classification settings, where specific groups may experience discrimination due to biased predictions. 
This has led to the formalization of several algorithmic fairness notions, such as \textit{Demographi Parity} \citep{Dwork2012}, \textit{Equal Opportunity} \citep{Hardt2016}, and \textit{Accuracy Parity} \citep{Zafar2017}. 
These notions aim to equalize various quantities across different groups.
While \textit{perfect fairness}--ensuring exactly identical quantities across groups--may entirely eliminate discrimination, it often incurs significant efficiency loss and may even be infeasible on finite data  \citep{agarwal2018reductions,makhlouf2021machine,Pinzon2022}. 
{Thus, \textit{approximate fairness} is frequently adopted as a more practical alternative, where fairness level is quantified and limited using approximate measures such as \textit{Mean Difference} \citep{chai2022fairness} and \textit{Mean Ratio} \citep{menon2018} derived from fairness notions. 
See Section \ref{sec:General_Fairness_Measures} for their definitions.}
 
Researchers have developed various fair ML algorithms to operationalize these fairness notions, which are typically categorized into three groups: pre-processing, in-processing, or post-processing \citep{caton2024fairness}.
 Pre-processing methods aim to reduce bias in the training data through techniques such as data cleaning or reweighting \citep{kamiran2012data,Calmon2017} before applying classical ML algorithms, but fairness in the training data does not always guarantee fairness in the resulting models.
In-processing methods modify the model training objective by adding fairness regularizers or incorporating  fairness constraints \citep{zemel2013learning,agarwal2018reductions,zafar2017fairnessMec,Yang2020,Zhao2020Conditional}.
Post-processing  \citep{menon2018,gouic2020projection,xian2023fair,chen2024posthoc,xian2024unifiedpostprocessing,wei2021optimized,cruz2024unprocessing} remaps the model’s outputs to satisfy fairness requirements.

Despite these advancements, foundational theoretical aspects of fair ML remain under-explored. 
One critical question concerns the characterization of Bayes-optimal classifiers in fair ML.
A Bayes-optimal fair classifier minimizes classification risk while satisfying specific fairness constraints, serving as a theoretical benchmark or the “best possible” classifier for a given fairness-aware problem.
Although \citet{menon2018} and \citet{chzhen2019leveraging} characterized Bayes-optimal fair classifiers, their analyses are limited to a single sensitive feature of binary values. 
This leaves more complex and realistic settings involving multiple sensitive features\footnote{Alternatively, a multi-class sensitive feature--see Section~\ref{sec:Fairness-aware_Learning_Binary_Classification} for details on their connection.} largely unaddressed.
While several studies \citep{Corbett2017,Schreuder2021,zeng2024bayesoptimalfair} have investigated the theoretical underpinnings of fair classification with multiple sensitive features, their works derive Bayes-optimal classifiers under the strict requirement of perfect fairness without addressing the practical requirement of approximate fairness.
More recently, \citet{chen2024posthoc} and \citet{xian2024unifiedpostprocessing} extended the exploration of Bayes-optimal fair classifiers to approximate fairness settings with multiple sensitive features.
However, their work focuses exclusively on post-processing algorithms for fair classification, and modifying model outputs in this manner may raise legal concerns \citep{caton2024fairness,Barocas2016}.
Their works are also restricted to the mean difference measure and fail to accommodate fairness notions like \textit{accuracy parity}. 
For a summary and comparison with related work, see Table \ref{tab:related_work_comparision}, and a more detailed discussion is provided in Appendix \ref{apxsec:Relatedwork} of the supplementary material. 

\begin{table*}[t]

\vskip 0.1in 
\begin{center}
\begin{small}
\resizebox{\textwidth}{!}{%
\begin{tabular}{lp{1.7cm}p{1.85cm}p{1.3cm}p{1.3cm}p{1.3cm}p{2cm}c}
\toprule
References & \citet{Corbett2017} & \citet{Schreuder2021} & \citet{agarwal2018reductions} & \citet{chen2024posthoc}& \citet{xian2024unifiedpostprocessing} & \citet{zeng2024bayesoptimalfair}& Ours\\
\midrule
\multicolumn{7}{l}{\textsc{Scope of Theoretical Framework}}\\
\hline
Approximate Fairness (MD) &  &  &{\checkmark}&{\checkmark}&{\checkmark}& &{\checkmark}\\
Approximate Fairness (MR) &  &  && &   & &{\checkmark} \\
Attribute-Blind Setting\footnotemark & &  & {\checkmark}&{\checkmark} & {\checkmark} &{\checkmark}&{\checkmark}\\ 
\hline
\multicolumn{7}{l}{\textsc{Fairness Metrics Considered}}\\
\hline
Demographic Parity & {\checkmark} & {\checkmark} &{\checkmark}&{\checkmark} & {\checkmark} & {\checkmark}&{\checkmark}\\ 
Equal Opportunity & {\checkmark} &  & &{\checkmark} & {\checkmark} & {\checkmark}&{\checkmark}\\ 
Predictive Equality  & {\checkmark} &  &  & &  & {\checkmark}&{\checkmark}\\ 
Accuracy Parity &  & &  &  &   &  &{\checkmark}\\ 
Equalized Odds & &  &{\checkmark}& {\checkmark} & {\checkmark} & {\checkmark} &{\checkmark}\\ 
\hline
\multicolumn{7}{l}{\textsc{Theoretically Optimal Algorithms}}\\
\hline
In-processing &  & &{\checkmark}&  &  & {\checkmark}&{\checkmark}\\
Post-processing & {\checkmark} & {\checkmark} & &{\checkmark} & {\checkmark} & {\checkmark}&{\checkmark}\\
\bottomrule
\end{tabular}
}
\end{small}
\end{center}
\caption{Our contributions in comparison with prior theoretical works for Bayes-optimal fair classifier with multiple sensitive features.}
\label{tab:related_work_comparision}
\vskip -0.1in
\end{table*}

\footnotetext{Attribute-Blind Setting refers to the case where sensitive features cannot be used for prediction.}

Therefore, it lacks a systematic approach for deriving Bayes-optimal fair classifiers, especially with multiple sensitive features under general approximate fairness measures. 
To this end, we explore their form while also explicitly addressing fairness notions such as accuracy parity.
{Our contributions are listed as follows:
\begin{itemize}
    \item  We characterize the form of Bayes-optimal fair classifiers for multiple sensitive features under both MD and MR measures, generalizing the framework of \citet{menon2018}.
    Their work can be viewed as a special case of our approach when restricted to a single (binary) sensitive feature. 
    { \item Our characterization accommodates fairness notions such as accuracy parity, whose Bayes-optimal fair classifier, to the best of our knowledge, has not been established before.
    \item Building on theoretical results, we propose both in-processing and post-processing algorithms to recover Bayes-optimal fair classifiers,  offering flexibility in when to apply fairness interventions.}
\end{itemize}
}

%% file: 2-Preliminaries.tex
\section{Background and Notation}\label{sec:Background&Notation}

\subsection{Binary Classification}
A binary classification problem is defined by a joint distribution $\mathcal{D}$ over input features $X \in \mathcal{X}$ and labels $Y \in \mathcal{Y} = \{0, 1\}$. 
The goal is to derive a measurable \textit{randomized classifier} parametrized by $f : \mathcal{X} \to [0, 1]$,  which outputs a prediction $\hat{Y}_{f} \in \{0, 1\}$ with a certain probability based on features $X$. 
{ Let $\mathrm{Bern}(p)$ be the Bernoulli distribution} with success probability $p \in [0, 1]$, and let $\mathcal{F}$ denote the set of all such measurable functions $f$. 
Then, the randomized classifier $f \in \mathcal{F}$ specifies, for any $x \in \mathcal{X}$, the probability $f(x)$ of predicting $\hat{Y}_{f} = 1$ given $X = x$, i.e., $\hat{Y}_{f} \mid X = x \sim \mathrm{Bern}(f(x))$. 

Typically, the quality of a classifier is evaluated using a statistical risk function $R(\cdot; \mathcal{D}) : \mathcal{F} \to \mathbb{R}_{+}$. 
A canonical risk is the cost-sensitive risk \citep{menon2018}.
\begin{definition}[Cost-Sensitive Risk]\label{df:CS_risk}
For a cost parameter $c \in [0, 1]$ and a classifier $f$,
the cost-sensitive risk (of $f$) is given by: 
\begin{align}\label{eq:CS_Risk_obj}
R_{cs}(f;c) =(1-c)\cdot P(\hat{Y}_{f}=0, Y=1) +c\cdot P(\hat{Y}_{f}=1, Y=0). 
\end{align}
\end{definition}
The cost-sensitive risk allows for asymmetric penalization of false negatives and false positives, depending on the value of $c$.
When $c = 0.5$, it reduces to the conventional error rate.

\textbf{Bayes-Optimal Classifiers:}
For a given problem, the Bayes-optimal classifier is theoretically the best method, achieving the lowest possible average risk.
For the cost-sensitive risk with parameter $c$, a Bayes-optimal classifier is defined as any minimizer $ f^{*} \in \argmin_{f\in\mathcal{F}} R_{cs}(f ; c)$ .
Let $\eta(x) \coloneqq P(Y = 1 \mid X = x)$  be the posterior probability of the positive class given $x$, and $\mathds{1}[\cdot]$ denote the indicator function (equal to $1$ if the argument is true and $0$ otherwise).
Then, \citet{elkan2001foundations} characterizes Bayes-optimal classifiers as having the form of 
\begin{equation}\label{eq:Bayesclassifier_S}
    f^{*}(x)=\mathds{1}\left[H(x)>0\right]+\alpha\cdot\mathds{1}\left[H(x)=0\right],
\end{equation}
for all $x\in\mathcal{X}$, where $H(x)=\eta(x)-c$, and $\alpha \in [0, 1]$ is an arbitrary parameter. This shows that the Bayes-optimal classifier operates as a \textit{thresholding rule} on the posterior class-probability of an instance. 
It makes predictions based on the threshold defined by the cost parameter $c$.

\subsection{Fairness-Aware Learning in Binary Classification}\label{sec:Fairness-aware_Learning_Binary_Classification}

Fairness-aware learning extends the conventional binary classification problem by incorporating sensitive features in addition to the target feature $Y$. Specifically, we assume the presence of sensitive features $A \in \mathcal{A}$ (e.g., gender and race) with respect to which we aim to ensure fairness. 
We note that ${X}$ may or may not include the sensitive features ${A}$ in practical applications.

\textbf{Group Notation with Multiple Sensitive Features}: In real applications, individuals might be coded with multiple sensitive features. 
We consider $K$ sensitive features, where each feature is denoted by $A_k \in \mathcal{A}_k$ for $k \in [K]$.\footnote{Here, $\mathcal{A} = \mathcal{A}_1 \times \mathcal{A}_2 \times \cdots \times \mathcal{A}_K$.} 
For example, $A_1$ might correspond to race, $A_2$ to gender, and so on.
However, the presence of multiple sensitive features (e.g., race and gender
simultaneously) can lead to non-equivalent definitions of group fairness \citep{Yang2020}:
{ \begin{itemize} [leftmargin=2em]
    \item \textit{Independent group fairness}: Fairness is evaluated separately for each sensitive feature, leading to overlapping subgroups (i.e., each sensitive feature defines its own set of groups independently).\footnote{See Appendix \ref{apxsec:Independent_Intersectional_Extensions} of the supplementary material for detailed explanations and examples.}
    \item \textit{Intersectional group fairness}: Fairness is enforced on all subgroups defined by intersections of sensitive features, resulting in non-overlapping groups associated with all possible combinations of sensitive features. 
\end{itemize}}

It is noteworthy that enforcing intersectional fairness inherently controls independent fairness, but the reverse does not always hold \citep{kearns2018preventing}.
Thus, intersectional fairness is often considered ideal \citep{Yang2020}.
Consequently, we focus on intersectional fairness here when addressing multiple sensitive features and also extend our results to independent fairness in Appendix \ref{apxsec:extension_indepent} of the supplementary material.

To implement intersectional fairness for multiple sensitive features, a new composite sensitive feature $S$ is constructed to represent all possible intersectional combinations of the existing sensitive features. Specifically, $S \in \mathcal{S}=\{1, \dots, M\}$, where 
$M = \prod_{k=1}^{K} |\mathcal{A}_k|$,
and $|\mathcal{A}_k|$ denotes the number of possible values for the $k$-th sensitive feature. 
Thus, $S$ defines $M$ non-overlapping subgroups, each corresponding to a unique combination of sensitive feature values.
Note that this approach is equivalent to treating $S$ as a single sensitive feature with multiple categorical values, {enabling our results to be directly applicable to that scenario}.

For all $m \in \mathcal{S}$, $x \in \mathcal{X}$, and $y \in \mathcal{Y}$, let $P_{S,Y}(m, y) := P(S = m, Y = y)$ denote the joint distribution of $S$ and $Y$.
Define $p^{+} := P(Y = 1)$ and $p^{-} := P(Y = 0)$ as the marginal probabilities of the positive and negative classes.
Let $P_{S}(\cdot)$ represent the marginal distribution of $S$, while $P_{S \mid Y = y}(\cdot)$ and $P_{Y \mid S = m}(\cdot)$ denote the conditional distributions of $S$ given $Y = y$ and $Y$ given $S = m$, respectively.

To address unfairness, various parity-based group fairness notions grounded in sensitive features have been proposed.
Below are the key definitions considered in this paper. 
\begin{definition}[Demographic Parity (DP)]\citep{Dwork2012}
A classifier $f$ satisfies DP if its prediction $\hat{Y}_{f}$ is independent of the sensitive feature $S$: 
$P(\hat{Y}_{f} = 1) = P(\hat{Y}_{f} = 1 \mid S = m)$
for all $m \in [M]$.
\end{definition}

\begin{definition}[Equal Opportunity (EO)]\citep{Hardt2016}
A classifier $f$ satisfies EO if it achieves the same true positive rate across all groups: $P(\hat{Y}_{f} = 1 \mid Y = 1) = P(\hat{Y}_{f} = 1 \mid S = m, Y = 1)$ for all $m \in [M]$.
\end{definition}

\begin{definition}[Predictive Equality (PE)]\citep{Corbett2017}
A classifier $f$ satisfies PE if it achieves the same false positive rate across all groups: $P(\hat{Y}_{f} = 1 \mid Y = 0) = P(\hat{Y}_{f} = 1 \mid S = m, Y = 0)$
for all $m\in [M]$.
\end{definition}

\begin{definition}[Accuracy Parity (AP)]\citep{Zafar2019}
A classifier $f$ satisfies AP if it achieves the same error rate across all groups: $P(\hat{Y}_{f} \neq Y) = P(\hat{Y}_{f} \neq Y \mid S = m)$
for all $m\in [M]$.
\end{definition}

{ In practice, perfect fairness (i.e., achieve equalities above) often leads to significant efficiency loss (i.e., higher expected risk) or even is infeasible \citep{makhlouf2021machine}.  
Thus,  ``approximate'' fairness is usually more practical and preferable.
Previous research typically quantifies fairness by measuring disparities in quantities that would be equalized under perfect fairness, focusing on optimizing risk while imposing constraints to limit these disparities \citep{Zafar2017}.}

%% file: 3-Genernal_fair_measure.tex
\section{General Approximate Fairness Measures}\label{sec:General_Fairness_Measures}
We focus on two general approximate fairness measures, \textit{mean difference}  and \textit{mean ratio}, to quantify classifier disparity level. 
We begin by presenting the definitions of these measures and then demonstrate both of them are linear transformations of a classifier's selection rates for specific groups.

\begin{table*}[!t]
\begin{center}
\resizebox{\textwidth}{!}{%
\begin{tabular}{p{0.5cm}cp{0.1cm}p{0.1cm}p{1.4cm}cccc}
\toprule
 Notion & $\mathcal{G}(\hat{Y}_{f})$ & $Z$ & $z$ & $a_{m}$ & $b_{m}^{y}$ & $c_{m}^{y}$ (MD) & $c_{m}^{y}$ (MR)\\
\midrule
DP    & $\{\hat{Y}_{f} = 1\}$ & $\mathbb{U}$\footnotemark & $\mathbb{U}$ & $P_{S}(m)$ & $P_{Y\mid S =m}(y)$ & $0$ & $0$
\\
EO    & $\{\hat{Y}_{f} = 1\}$ & $Y$ & $1$ & $P_{S\mid Y =1}(m)$ & $y$ & $0$ & $0$ \\
PE    & $\{\hat{Y}_{f} = 1\}$ & $Y$ & $0$ & $P_{S\mid Y =0}(m)$ & $1 - y$ & $0$ & $0$ \\
{AP} & {$\{\hat{Y}_{f} \neq Y\}$} & {$\mathbb{U}$} & {$\mathbb{U}$} & {$P_{S}(m)$} & {$(1 - 2y) \cdot P_{Y\mid S =m}(y)$} & $(1-y)p^{+}-yP_{Y\mid S =m}(1)$ & {$(y-1)\delta p^{+}+yP_{Y\mid S =m}(1)$} \\
\bottomrule
\end{tabular}
}
\end{center}
\caption{Recovering existing fairness criteria based on the choice of $ \mathcal{G}(\cdot) $, $Z$, and $z$ for MD and MR measures. 
For parameter values in Lemmas \ref{lem:MD_FNR_risk} and \ref{lem:MR_FNR_risk},  MD and MR measures differ only in $c_m^{y}$ for AP.}
\label{tab:genernal_fair_measure}
\end{table*}

\subsection{Mean Difference}

For a composite sensitive feature $S \in \{1, \dots, M\}$, the \textit{mean difference} (MD) score \citep{ chai2022fairness,calders2010three} quantifies the fairness of a classifier $f$ by calculating the difference in a specified outcome between the overall population and the subgroup defined by $S = m$.

\begin{definition}[Mean Difference]\label{df:mean_diff}
For $\forall m\in [M]$, the mean difference measure for group $m$ is defined as:
\begin{align}\nonumber
    \mathrm{MD}_{m}(f) = P(\mathcal{G}(\hat{Y}_{f}) \mid Z=z) 
   - P(\mathcal{G}(\hat{Y}_{f}) \mid Z=z, S=m),
\end{align}
where $ \hat{Y}_{f} $ is the prediction of $ f $, and the components $ \mathcal{G}(\cdot) $, $Z$,  and $z$ depend on the fairness notion being considered. 
\end{definition}

\footnotetext{{$\mathbb{U}$} refers to the complete set.}

The flexibility in the choice of $ \mathcal{G}(\cdot) $, $Z$,  and $z$ allows Definition \ref{df:mean_diff} to accommodate several commonly used group fairness notions, as shown in Table \ref{tab:genernal_fair_measure}.
Achieving perfect fairness indicates $\mathrm{MD}_{m}(f) = 0$ for all $m$. 
Usually, a limited level of disparity may be acceptable.
To formalize this, we use the symmetrized version of the MD measure:
\begin{equation}\label{eq:mean_diff_constr}
\hspace{-0.06in}\mathrm{MD}(f)=\max_{m\in [M]} \max\left(\mathrm{MD}_{m}(f), \mathrm{MD}_{m}(1-f)\right) \leq \delta,
\end{equation}
where $\delta$ is a pre-specified tolerance level for unfairness.

To simplify notation, we define $E_{y,m}=\{Y=y,S=m\}$ as the event where an individual has label $Y=y$ and belongs to group $m$, with its probability denoted by $P(E_{y,m})=P(Y=y,S=m)$.
Then, Lemma \ref{lem:MD_FNR_risk} shows that MD measures for these common group fairness notions are linear transformations of $P(\hat{Y}_{f}=1\mid E_{y,m})$. 
All proofs are deferred to Appendix \ref{apxsec:Proofs} of the supplementary material.
\begin{lemma}\label{lem:MD_FNR_risk}
For any randomized classifier $f$, any $\delta\in[0,1]$, and the group fairness notions in Table \ref{tab:genernal_fair_measure},
$\mathrm{MD}(f)\leq \delta \Leftrightarrow R^\mathrm{MD}_{m}(f) \in [-\delta, \delta]$ for all $m\in [M]$,
where
\begin{align}\nonumber
R^\mathrm{MD}_{m}(f)  :=\sum_{y\in\{0,1\}} \Bigg\{\Big[
\sum_{m'=1}^M a_{m'} b^{y}_{m'} P(\hat{Y}_{f}=1\mid E_{y,m'})
\Big] 
-b_m^y  P(\hat{Y}_{f}=1\mid E_{y,m})
+ c_m^y
\Bigg\}. 
\end{align}
Here, values of $a_{m}$, $b_{m}^{y}$, and $c_{m}^{y}$ depend on the chosen fairness notion and are as defined in Table \ref{tab:genernal_fair_measure}.
\end{lemma}

\subsection{Mean Ratio}

Approximate fairness can also be assessed using the \textit{disparate impact} factor \citep{feldman2015certifying,menon2018}, which is defined as the ratio of relevant probabilities.  
We refer to this as the \textit{mean ratio} (MR) measure, shown below.
\begin{definition}[Mean Ratio]\label{df:mean_ratio}
For  $\forall m\in [M]$, the mean ratio measure for group $m$ is defined as:
\begin{equation}\nohao
   \mathrm{MR}_{m}(f) = \frac{P(\mathcal{G}(\hat{Y}_{f}) \mid Z=z, S = m)}{P(\mathcal{G}(\hat{Y}_{f}) \mid Z=z)},
\end{equation}
where $\hat{Y}_{f}$, $\mathcal{G}(\cdot)$, $Z$, and $z$ are as defined in Definition~\ref{df:mean_diff}.
\end{definition}
Similarly, we consider the symmetrized version of the MR measure \citep{menon2018}:
\begin{equation}\label{eq:mean_ratio_constr} 
\hspace{-0.1in}\mathrm{MR}(f)=\min_{m\in [M]} \min\left(\mathrm{MR}_{m}(f), \mathrm{MR}_{m}(1-f)\right) \geq \delta.  
\end{equation}
Then, Lemma \ref{lem:MR_FNR_risk} demonstrates that MR measures for common group fairness notions are also linear transformations of $P(\hat{Y}_{f}=1\mid E_{y,m})$.

\begin{lemma}\label{lem:MR_FNR_risk}
    For any randomized classifier $f$, any $\delta\in[0,1]$, and the group  fairness notions in Table \ref{tab:genernal_fair_measure}, $\mathrm{MR}(f)\geq \delta \Leftrightarrow R^\mathrm{MR}_{m}(f)\in[\delta-1,0]$ for all $m\in [M]$, where
\begin{align}\nohao
R^\mathrm{MR}_{m}(f):=\sum_{y\in\{0,1\}} \Bigg\{\Big[\delta
\sum_{m'=1}^M a_{m'} b^{y}_{m'} P(\hat{Y}_{f}=1\mid E_{y,m'})
\Big] 
-b_m^y  P(\hat{Y}_{f}=1\mid E_{y,m})
+ c_m^y
\Bigg\}. 
\end{align}
Here, values of $a_{m}$, $b_{m}^{y}$, and $c_{m}^{y}$ depend on the chosen fairness notion and are as defined in Table \ref{tab:genernal_fair_measure}.  
\end{lemma}

%% file: 4-Bayes-optimal_fair_classifier.tex
\section{Bayes-Optimal Fair  Classifiers}\label{sec:Fair_Bayes-Optimal_Classifiers}

Given fairness constraints in  \eqref{eq:mean_diff_constr} or \eqref{eq:mean_ratio_constr}, our goal is to find  a (randomized) fair classifier $f^{*}_{B}$ optimizing the following problems:
$\min_{f \in \mathcal{F}} \left\{R_{cs}(f; c) : \mathrm{MD}(f) \leq \delta\right\}$ for MD, or $\min_{f \in \mathcal{F}} \left\{R_{cs}(f; c) : \mathrm{MR}(f) \geq \delta\right\}$ for MR.
Note that these constrained optimization problems can be further reduced to the following unconstrained problems via the Lagrangian principle and Lemmas \ref{lem:MD_FNR_risk} and \ref{lem:MR_FNR_risk}.
\begin{lemma}\label{lem:Lagrangian_formlation} 
For any $c \in [0,1]$ and $\delta \in [0,1]$, there exists $\bs{\lambda} \in \mathds{R}^{M}$ such that:
\begin{itemize}[leftmargin=2em,itemsep=0pt, parsep=0pt, topsep=0pt]
    \item For MD:
    $\min_{f \in \mathcal{F}} \left\{R_{cs}(f; c) : \mathrm{MD}(f) \leq \delta\right\} = \min_{f \in \mathcal{F}} \left(R_{cs}(f; c) - \sum_{m=1}^{M} \lambda_m \cdot R^\mathrm{MD}_{m}(f)\right).$
    \item For MR:
    $\min_{f \in \mathcal{F}} \left\{R_{cs}(f; c) : \mathrm{MR}(f) \geq \delta\right\} = \min_{f \in \mathcal{F}} \left(R_{cs}(f; c) - \sum_{m=1}^{M} \lambda_m \cdot R^\mathrm{MR}_{m}(f)\right).$
\end{itemize}
Here, $\lambda_{m}$ is the $m$-th component of $\bs{\lambda}$.
\end{lemma}

Lemma \ref{lem:Lagrangian_formlation} shows that Bayes-optimal fair  classifiers can be derived by solving an unconstrained optimization problem with a fairness regularizer incorporated into the objective.
The trade-off parameter vector $\bs{\lambda}$ controls the balance between cost-sensitive risk (efficiency) and fairness. 
In fact, each of its component $\lambda_{m} \in \mathds{R}$ corresponds to the difference in Lagrange multipliers for the two bounds associated with group $m$, and it can take negative values.
With these foundations in place, we now present the form of Bayes-optimal fair classifiers for MD and MR measures.

\subsection{Mean Difference}
We begin with the explicit form of the Bayes-optimal fair classifier for MD measure.
Recall that $\eta(x):=P(Y=1\mid X=x)$.
\begin{theorem}[Bayes-Optimal Fair Classifier for MD]\label{th:Fairbayes_form_MD}
For any $c\in[0,1]$ and $\delta\in[0,1]$, $\exists \bs{\lambda}\in\mathds{R}^{M}$ such that the Bayes-optimal fair classifier  
$f_{B}^{*}(x)\in\argmin_{f\in\mathcal{F}}\left\{R(f): \mathrm{MD}(f)\leq\delta\right\}$
has the form of 
\begin{equation}\label{eq:bayes_classifier_MD}
f_{B}^{*} (x)=\mathds{1}\left[H_{B}^{*}(x)>0\right]+\alpha\cdot\mathds{1}\left[H_{B}^{*}(x)=0\right],
\end{equation}
where 
\begin{equation}\nohao
H_{B}^{*}(x)=\eta(x)-c-
\sum_{m=1}^M \sum_{y \in \{0,1\}}
    \hspace{-1pt} b_m^y\left(\lambda_m -\Lambda_M a_m\right)\gamma_m^y(x).    
\end{equation}
Here, $\lambda_{m}$ is the $m$-th component of $\bs{\lambda}$, $\Lambda_{M}=\sum_{i=1}^{M}\lambda_{m}$,  
$\gamma_{m}^{y}(x)=\frac{P(E_{y,m}\mid X=x)}{{P\left(E_{y,m}\right)}}$, and $\alpha \in [0, 1]$ is an arbitrary parameter. 
The values of $a_{m}$ and $b_{m}^{y}$ depend on the fairness notion under consideration  and are as shown in Table \ref{tab:genernal_fair_measure}.
\end{theorem}
In \eqref{eq:bayes_classifier_MD}, setting $\bs{\lambda} = \bs{0}$ results in the unconstrained Bayes-optimal classifiers for the cost-sensitive risk, as described in \eqref{eq:Bayesclassifier_S}. 
For $\bs{\lambda} \neq \bs{0}$, the optimal classifier $f_{B}^{*}(x)$ adjusts the $\bs{\lambda} = \bs{0}$ solution by applying an instance-dependent threshold correction. 
This correction is determined by the weighted sum of $\gamma_{m}^{y}(x)$—the (normalized) probability that the individual $x$ belongs to the group $\{Y=y, S=m\}$. 

In the discussion above, we made no explicit assumption regarding whether the sensitive features are utilized during the prediction phase.
Thus, the findings are applicable to the attribute-blind setting.
If the sensitive features are available and allowed to be used for prediction,\footnote{This refers to the Attribute-Aware Setting, i.e., $A$ (and thus $S$) are included in $X$.
In what follows, we slightly abuse notation by separating $A$ ($S$) from $X$, with $X$ representing only the non-sensitive features.} the form of the Bayes-optimal fair classifier simplifies as follows:
\begin{corollary}
[Bayes-Optimal Fair Classifier for MD-$S$]\label{corl:BayesClass_withS_MD}
For any $c\in[0,1]$ and $\delta\in[0,1]$, $\exists \bs{\lambda}\in\mathds{R}^{M}$ such that the Bayes-optimal fair classifier  
$f_{B}^{*}(x,s)\in\argmin_{f\in\mathcal{F}}\left\{R(f): \mathrm{MD}(f)\leq\delta\right\}$ 
has the form of 
\begin{equation}\label{eq:bayes_classifier_MD_S}
f_{B}^{*} (x,s)=\mathds{1}\left[H_{B}^{*}(x,s)>0\right]+\alpha\cdot\mathds{1}\left[H_{B}^{*}(x,s)=0\right],   
\end{equation}
where 
\begin{equation}\nohao
    H_{B}^{*}(x,s)=\eta(x,s)-c-\sum_{y\in\{0,1\}}b_{s}^{y}\left(\lambda_{s}-\Lambda_{M}a_{s}\right){\gamma_{s}^{y}(x,s)}.
\end{equation}
Here, $\eta(x,s)=P(Y=1\mid X=x, S=s)$, $\lambda_{s}$ is the $s$-th component of $\bs{\lambda}$, $\Lambda_{M}=\sum_{i=1}^{M}\lambda_{m}$, 
and $\gamma_{s}^{y}(x,s)=\frac{P(Y=y\mid X=x, S=s)}{P\left(E_{y,s}\right)}$. 
$\alpha \in [0, 1]$ is an arbitrary parameter. 
The values of $a_{s}$ and $b_{s}^{y}$ depend on the selected fairness notion.
\end{corollary}
This result follows directly from Theorem~\ref{th:Fairbayes_form_MD}, since for the data pair $(x, s)$, we have 
$P(S = m \mid X = x, S = s) = \mathds{1}[m = s]$ and 
$P(S = m, Y \mid X = x, S = s) = P(Y \mid X = x, S = s)\mathds{1}[m = s].$ 
Note that in this case, \eqref{eq:bayes_classifier_MD_S} can further reduce to applying a group-wise constant threshold to the class probabilities $\eta(x, s)$ for each value of the sensitive feature. 
This simplification arises because $\gamma_{s}^{y}(x, s)$ is a linear function 
of $\eta(x, s)$ across all four fairness notions.

\subsection{Mean Ratio}
We now turn to the Bayes-optimal fair classifier  for the MR measure. 
The result is analogous to Theorem \ref{th:Fairbayes_form_MD}, but it explicitly incorporates $\delta$ in the threshold correction.
\begin{theorem}[Bayes-Optimal Fair Classifier for MR]\label{th:Fairbayes_form_MR}
For any $c\in[0,1]$ and $\delta\in[0,1]$, $\exists \bs{\lambda}\in\mathds{R}^{M}$ such that the Bayes-optimal fair classifier  
$f_{B}^{*}(x)\in\argmin_{f\in\mathcal{F}}\left\{R(f): \mathrm{MR}(f)\geq\delta\right\}$
has the form of
\begin{equation}\label{eq:bayes_classifier_MR}
f_{B}^{*} (x)=\mathds{1}\left[H_{B}^{*}(x)>0\right]+\alpha\cdot\mathds{1}\left[H_{B}^{*}(x)=0\right],
\end{equation}
where 
\begin{equation}\nohao
H_{B}^{*}(x)=\eta(x)-c-\hspace{-1pt}
\sum_{m=1}^M \sum_{y \in \{0,1\}}
    \hspace{-5pt}b_m^y\left(\lambda_m\hspace{-2pt}-\delta\hspace{-1pt}\cdot\hspace{-1pt}\Lambda_M a_m\right)\gamma_m^y(x).
\end{equation}
Here, $\lambda_{m}$ is the $m$-th component of $\bs{\lambda}$, $\Lambda_{M}=\sum_{i=1}^{M}\lambda_{m}$,  
$\gamma_{m}^{y}(x)=\frac{P(E_{y,m}\mid X=x)}{{P\left(E_{y,m}\right)}}$, and $\alpha \in [0, 1]$ is an arbitrary parameter. 
The values of $a_{m}$ and $b_{m}^{y}$ depend on the fairness notion under consideration and are as shown in Table \ref{tab:genernal_fair_measure}.
\end{theorem}    
When $S$ is available for prediction, the Bayes-optimal fair classifier for MR, similar to the MD case, takes a simplified form as detailed in Corollary~\ref{corl:BayesClass_withS_MR}.
\begin{corollary}[Bayes-Optimal Fair Classifier for MR-$S$]\label{corl:BayesClass_withS_MR}
 For any $c\in[0,1]$ and $\delta\in[0,1]$, $\exists \bs{\lambda}\in\mathds{R}^{M}$ such that the Bayes-optimal fair classifier $f_{B}^{*}(x,s)\in\argmin_{f\in\mathcal{F}}\left\{R(f): \mathrm{MR}(f)\geq\delta\right\}$
has the form of
\begin{equation}\label{eq:bayes_classifier_MR_S}
f_{B}^{*} (x,s)=\mathds{1}\left[H_{B}^{*}(x,s)>0\right]+\alpha\cdot\mathds{1}\left[H_{B}^{*}(x,s)=0\right],   
\end{equation}
where 
\begin{equation}\nohao
H_{B}^{*}(x,s)=\eta(x,s)-c-\hspace{-5pt}\sum_{y\in\{0,1\}}b_{s}^{y}\left(\lambda_{s}-\delta\cdot\Lambda_{M}a_{s}\right){\gamma_{s}^{y}(x,s)}.
\end{equation}
Here, $\eta(x,s)=P(Y=1\mid X=x, S=s)$, $\lambda_{s}$ is the $s$-th component of $\bs{\lambda}$, $\Lambda_{M}=\sum_{i=1}^{M}\lambda_{m}$, 
and $\gamma_{s}^{y}(x,s)=\frac{P(Y=y\mid X=x, S=s)}{P\left(E_{y,s}\right)}$. 
$\alpha \in [0, 1]$ is an arbitrary parameter. 
The values of $a_{s}$ and $b_{s}^{y}$ depend on the selected fairness notion.
\end{corollary}
Similar to the MD case, \eqref{eq:bayes_classifier_MR_S} can further reduce to applying a group-wise constant threshold to the class probabilities $\eta(x, s)$ for each value of the sensitive feature.
\begin{remark}
In addition to fairness notions discussed in Table \ref{tab:genernal_fair_measure}, our results can be directly extended to composite fairness notions like \textit{Equalized Odds} \citep{Hardt2016}. 
See Appendix \ref{apxsec:extension_composite} of the supplementary material for more details.
\end{remark}

%% file: 5-Algorithm.tex
\section{Algorithms}\label{sec:Algorithms}
Section \ref{sec:Fair_Bayes-Optimal_Classifiers} establishes that Bayes-optimal fair classifiers for MD and MR measures apply instance-dependent threshold corrections to the unconstrained Bayes-optimal classifier. 
This insight facilitates practical training on finite data using in-processing and post-processing techniques.    

\subsection{In-Processing-Based Bayes-Optimal Fair Classification} 
We first introduce an in-processing method. 
Theorems~\ref{th:Fairbayes_form_MD} and~\ref{th:Fairbayes_form_MR} show  that Bayes-optimal fair classifiers apply instance-dependent threshold adjustments. 
As cost-sensitive classification inherently accounts for such threshold adjustments, it forms the basis of our approach. 
We propose a fair cost-sensitive classification framework by first defining a fair cost-sensitive risk function and then demonstrating that minimizing this risk yields the Bayes-optimal fair classifier, as shown in the following theorem.
\begin{theorem}
\label{th:FairCS_BayesOptimal}
Let $ c^{\bs{\lambda}}_{y}(x) = (1 - 2y)\left[c + Q^{\bs{\lambda}}(x)\right] + y, $ where $ y \in \{0, 1\} $, and:
\begin{align}\label{eq:algo1_Q_MD}
    Q_\mathrm{MD}^{\bs{\lambda}}(x) &= \sum_{m=1}^M \sum_{y \in \{0,1\}}
    b_m^y\left(\lambda_m -\Lambda_M a_m\right)\gamma_m^y(x), \\\label{eq:algo1_Q_MR}
    Q_\mathrm{MR}^{\bs{\lambda}}(x) &= \sum_{m=1}^M \sum_{y \in \{0,1\}}
    b_m^y\left(\lambda_m -\delta\cdot\Lambda_M a_m\right)\gamma_m^y(x).
\end{align}
Here, $\bs{\lambda}$, $ \lambda_{m} $, $ \Lambda_{M}$, 
$ \gamma_{m}^{y}(x)$, and the values of $a_m$ and $b_m^y$ are as defined in Theorems~\ref{th:Fairbayes_form_MD}.

Define the fair cost-sensitive risk of a classifier $ f $ as:
\begin{align}\label{eq:fair_CS_risk}
    R_{FCS}^{\bs{\lambda}} (f) 
    =\sum_{y \in \{0,1\}} \Big\{\int_{\mathcal{X}} c^{\bs{\lambda}}_{y}(x) \cdot  P(\hat{Y}_{f} = 1-y, Y = y \mid X = x) \, dP_{X}(x)\Big\}.
\end{align}
Then, $f_{B}^\text{In}(x) = \argmin_{f \in \mathcal{F}} R_{FCS}^{\bs{\lambda}}(f)$ is a Bayes-optimal fair classifier.
\end{theorem}
Theorem \ref{th:FairCS_BayesOptimal} shows that a cost-sensitive classification with instance-dependent costs can yield a Bayes-optimal fair classifier.
Building on this, Algorithm~\ref{algo:in_processing} outlines  the proposed in-processing procedure.\footnote{We assume that $S$ has already been constructed from $A$, and so have its observed values $s_i$.}
\begin{algorithm}[ht]
\caption{Bayes-Optimal Fair Classification via In-Processing}\label{algo:in_processing}
    \begin{algorithmic}[1]
\renewcommand{\algorithmicrequire}{\textbf{Input:}}
\renewcommand{\algorithmicensure}{\textbf{Output:}}
  \REQUIRE Cost parameter $c$, fairness tolerance level $\delta\geq0$, dataset $D=\{x_i,s_i,y_i\}_{i=1}^{N}$, and trade-off parameter $\bs{\lambda}$.     
     \STATE \label{algo:algo1_line1} Estimate the group membership probabilities {$\tilde{P}(S, Y \mid X = x)$} using $\{(x_i, s_i, y_i)\}_{i=1}^{N}$, and calculate $\hat{\gamma}_{m}^{y}(x)=\frac{\tilde{P}(E_{y,m}\mid X=x)}{\tilde{P}(E_{y,m})}$ accordingly for all $m \in [M]$ and $y\in\{0,1\}$.
     \STATE Estimate $\hat{Q}^{\bs{\lambda}}$ by plug-in estimation using $\hat{\gamma}_{m}^{y}$ and the expressions in \eqref{eq:algo1_Q_MD} for MD or \eqref{eq:algo1_Q_MR} for MR.   
        \STATE Denote $ \hat{c}^{\bs{\lambda}}_{y}= (1 - 2y)\left[c + \hat{Q}^{\bs{\lambda}}\right] + y $ for all $y$. 
        \STATE Use any cost-sensitive classification method to train $\hat{f}_{B}^{*}(x)$ on $\{x_i,y_i\}_{i=1}^{N}$ by minimizing the empirical analogue of fair cost-sensitive risk in \eqref{eq:fair_CS_risk}. 
        \ENSURE $\hat{f_{B}^{*}}(x)$.
\end{algorithmic}
\end{algorithm}

\begin{remark}\label{rmk:group_pro_estimation}
On the line \ref{algo:algo1_line1} of Algorithm \ref{algo:in_processing}, the group membership probabilities are estimated.
This can be done directly by learning a multi-class predictor $\hat{f}_{S,Y}:\mathcal{X}\rightarrow \Delta^{\mathcal{S}\times\mathcal{Y}}$.
Or one can break the problem down into learning two simple predictors, $\hat{f}_Y: \mathcal{S}\times\mathcal{X}\rightarrow\Delta^\mathcal{Y}$ and  $\hat{f}_{S}:\mathcal{X}\rightarrow \Delta^{\mathcal{S}}$, then combining them into $\hat{f}_{S,Y}(m,y)=\hat{f}_{S}(x)_{m}\cdot\hat{f}_{Y}(m,x)_y$.  
\end{remark}    
When $S$ is available for prediction, the construction of the Bayes-optimal fair cost-sensitive classifier can be further simplified. 
See Appendix \ref{apxsec:C.8} of the supplementary material for details.

\subsection{Post-Processing-Based Bayes-Optimal Fair Classification}
Recall that Theorems \ref{th:Fairbayes_form_MD} and \ref{th:Fairbayes_form_MR} show that the Bayes-optimal fair classifiers $ f_{B}^{*}(x) $ modify the unconstrained  Bayes-optimal classifier by applying an instance-dependent threshold correction.
This correction depends on the (normalized) group membership probabilities of the individual $ x $, given by $\gamma_{m}^{y}(x)$.
Motivated by this, we propose a post-processing algorithm that adopts a plugin approach for Bayes-optimal fair classification. 
Specifically, we construct a fair plugin classifier by separately estimating $ \eta(x) $ and $\gamma_{m}^{y}(x)$, and then combining them according to \eqref{eq:bayes_classifier_MD} and~\eqref{eq:bayes_classifier_MR}.
{ See Appendix \ref{apxsec: Calibration} for  discussion regarding estimation probability calibration.}

Algorithm~\ref{algo:post_processing} outlines the proposed post-processing plug-in approach.
When $ S $ is available for prediction, we can estimate $ \hat{\eta}(x, s) = \Tilde{P}(Y = 1 \mid X = x, S = s) $ using any method applied to the dataset $ \{(x_i, s_i, y_i)\}_{i=1}^{N} $ instead of $ \{(x_i, y_i)\}_{i=1}^{N} $ as described in Line 1 of Algorithm~\ref{algo:post_processing}.

\begin{algorithm}[htb]
\caption{Bayes-Optimal Fair Classification via Post-Processing}\label{algo:post_processing}
    \begin{algorithmic}[1]
\renewcommand{\algorithmicrequire}{\textbf{Input:}}
\renewcommand{\algorithmicensure}{\textbf{Output:}}
  \REQUIRE Cost parameter $c$, fairness tolerance level $\delta\geq0$, dataset $D=\{x_i,s_i,y_i\}_{i=1}^{N}$, and trade-off parameter $\bs{\lambda}$.        
    \STATE\label{algo:algo2_line1} Estimate $\hat{\eta}(x) = \Tilde{P}(Y = 1 \mid X = x)$ using any approach on $\{(x_i, y_i)\}_{i=1}^{N}$.
        \STATE Estimate the group membership probabilities $\tilde{P}(S, Y \mid X = x)$ using $\{(x_i, s_i, y_i)\}_{i=1}^{N}$, and calculate $\hat{\gamma}_{m}^{y}(x)=\frac{\tilde{P}(E_{y,m}\mid X=x)}{\tilde{P}(E_{y,m})}$ accordingly for all $m \in [M]$ and $y\in\{0,1\}$.     
        \STATE Construct $\hat{f}_{B}^{*}(x)$ by plugging the estimates $\hat{\eta}(x)$ and $\hat{\gamma}_{m}^{y}(x)$ into the expression for the Bayes-optimal fair classifier: Use  \eqref{eq:bayes_classifier_MD} for MD or  \eqref{eq:bayes_classifier_MR} for MR.
        \ENSURE $\hat{f_{B}^{*}}(x)$.
\end{algorithmic}
\end{algorithm}

\begin{remark}\label{rmk:lambda}
In Algorithms \ref{algo:in_processing} and \ref{algo:post_processing}, the value of $\bs{\lambda}$ plays a critical role in balancing fairness and risk. 
It can be selected through various approaches. 
For example, a rough estimate can be obtained through grid search by ensuring that the achieved fairness level satisfies the pre-specified threshold \citep{menon2018,chen2024posthoc}.  
For a more accurate or efficient determination, optimization techniques based on the relationship between $\bs{\lambda}$ and the Lagrange multipliers can be employed, like solving a dual optimization problem \citep{xian2024unifiedpostprocessing} or using iterative updates guided by the fairness-accuracy trade-off \citep{zeng2024bayesoptimalfair}. 
{See Appendix \ref{apxsec:lambda_selection} for more details on the dual update procedure.}
\end{remark}

\begin{remark}\label{rmk:sparsity}

Our algorithms are  estimator-agnostic, existing sparsity  or calibration remedies can plug in directly when estimating group membership probability.
When $K$ is large, intersectional sparsity may occurs; 
see Appendix \ref{apxsec:Independent_Intersectional_Extensions} for discussion and its fixes. 
For many groups/high-dim $X$, scalable estimators \citep{zhang2014multicategory,liu2018smac} can help for efficiency.    
\end{remark}

%% file: 6-Experiments.tex
\section{Experiments}\label{sec:Experiments}

\textbf{Setup.} { We consider two real-world benchmark classification datasets that are widely used in the fair ML literature:  \texttt{Adult} \citep{adult_2} 
 and \texttt{COMPAS} \citep{angwin2016machinebias}. 
 Both datasets contain demographic features such as \textit{gender} and \textit{race}. 
We use them to construct scenarios with a single sensitive feature as well as multiple sensitive features.
We compare the proposed algorithms with established post-processing and in-processing fair classification algorithms, including \textit{LinearPost} \citep{xian2024unifiedpostprocessing}, {\textit{MBS} \citep{chen2024posthoc}}, and \textit{Reduction} \citep{agarwal2018reductions}. 
We also include a basic baseline classifier without fairness constraints for comparison, including LR and XGBoost. 
We consider both attribute-blind and attribute-aware settings. 
Following \citet{menon2018} and \citet{chen2024posthoc}, the values of the hyperparameter $ \bs{\lambda} \in  [-1,1]^{M}$ are selected via grid search in our algorithms. 
The cost parameter is fixed at $c = 0.5$. 
Detailed dataset statistics and model setups are provided in the Appendix \ref{apxsex:Experiments} of the supplementary material.}

\begin{figure*}[!htbp]
 \centering
 \includegraphics[width=\textwidth]{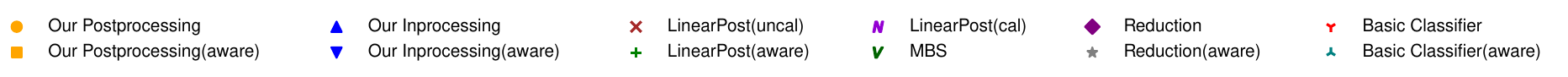} 
\begin{center} 
   \begin{subfigure}{0.49\textwidth}
        \centering
        \begin{subfigure}{0.49\textwidth} 
            \centering
            \includegraphics[width=\linewidth]{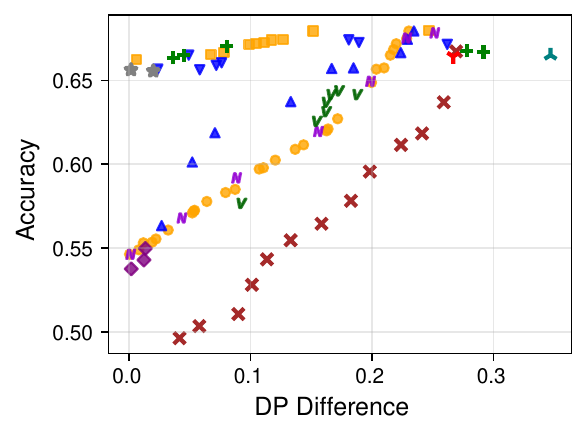}
        \end{subfigure}
        \begin{subfigure}{0.49\textwidth} 
            \centering
            \includegraphics[width=\linewidth]{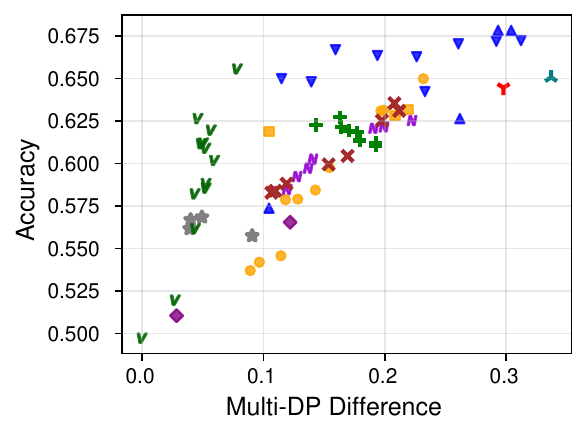}
        \end{subfigure}
        \caption{\texttt{COMPAS}}
        \label{fig:compas}
    \end{subfigure}
    \hfill
    \begin{subfigure}{0.49\textwidth}
        \centering
        \begin{subfigure}{0.49\textwidth} 
            \centering
            \includegraphics[width=\linewidth]{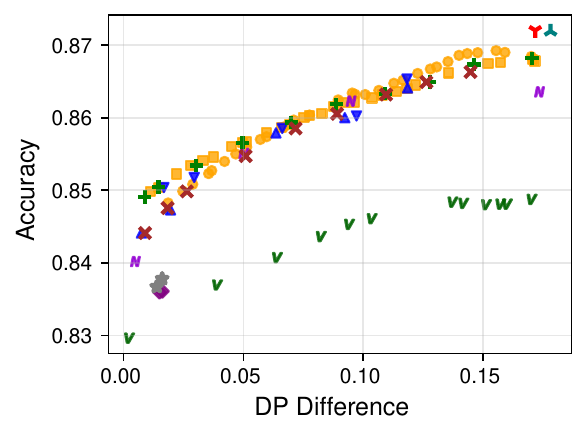}
        \end{subfigure}
        \begin{subfigure}{0.49\textwidth} 
            \centering
            \includegraphics[width=\linewidth]{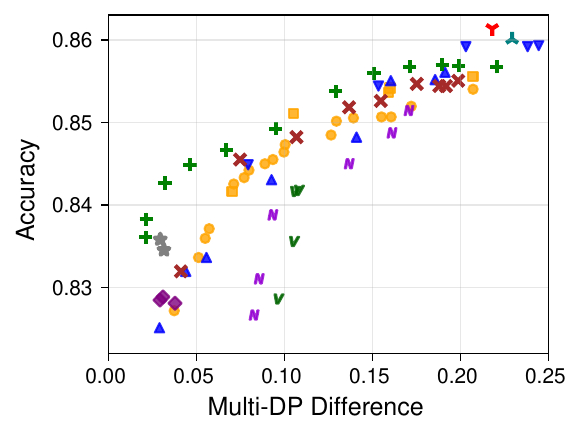}
        \end{subfigure}
        \caption{\texttt{Adult}}
        \label{fig:adult}
    \end{subfigure}
            
    \caption{Trade-offs between accuracy and fairness using MD. 
    The prefix `Multi-' represents the case of multiple sensitive features, and `(aware)' in classifier  names indicates the attribute-aware setting. (uncal): uncalibrated; (cal): calibrated.}
\label{fig:combined_data_fairness}
    \end{center}
  
\end{figure*}

{ \textbf{Evaluation Metrics.} For various values of $\bs{\lambda}$, we report both accuracy and fairness levels, as they represent the key performance metrics of interest. 
All four fairness notions in Table~\ref{tab:genernal_fair_measure} are considered. 
Approximate fairness is implemented using both MD and MR measures, and the achieved fairness level (i.e., values of $\mathrm{MD}(f)$ and $\mathrm{MR}(f)$) are reported.}
\textbf{Results.} 
{Figure \ref{fig:combined_data_fairness}} shows   fairness-accuracy trade-offs for MD measure under DP across two datasets. 
It considers cases where \textit{gender} or \textit{race} is the only sensitive feature and where both of them are considered sensitive. 
Each point corresponds to a specific value of the tuning parameter $\bs{\lambda}$.
Compared to the fair baselines, our methods achieve the more favorable fairness-accuracy trade-off in most cases, especially in the attribute-blind setting. 
Additionally, our in-processing method often outperforms our post-processing one in balancing fairness and accuracy, with a more significant advantage on the \texttt{COMPAS} dataset.
We also evaluate our methods under other fairness notions and with respect to the MR measure.
{Due to space constraints, the complete results are reported in Appendix \ref{apx:expresults} of the supplementary material, and Appendix \ref{apxsex:Additional_Experiments} presents additional sensitivity and ablation analyses.}
All results suggest that our methods effectively navigates the trade-off between fairness and accuracy.

%% file: 7-Conclusion.tex
\section{Conclusion}

This work analyzes the fair classification problem with multiple sensitive features. 
We characterize that Bayes-optimal fair classifiers for approximate fairness--under both \textit{mean difference} and \textit{mean ratio} measures--can be represented by instance-dependent threshold corrections applied to the unconstrained Bayes-optimal classifier. 
The  corrections are determined by a weighted sum of the probabilities that an individual belongs to specific groups.
Our findings are applicable to both attribute-aware and attribute-blind settings and cover widely used fairness notions, including DP, EO, PE, and \textit{accuracy parity} (AP).
Notably, to the best of our knowledge, this is the first work to characterize the Bayes-optimal fair classifier under AP.
Building on these insights, we proposed both in-processing and post-processing algorithms for learning Bayes-optimal fair classifiers from finite data.
Empirical results show that our methods perform favorably compared to existing methods.

%% file: 8-Appendix.tex

\section{Related Work}\label{apxsec:Relatedwork}
{
There are several works focusing on Bayes-optimal classifiers in fair ML in recent years.
For example, \citet{menon2018} and \citet{chzhen2019leveraging} characterized Bayes-optimal fair classifiers. 
However, their analyses are limited to settings with a single (binary) sensitive feature, leaving more complex and realistic scenarios involving multiple sensitive features largely unaddressed.
While several studies \citep{Corbett2017,Schreuder2021,zeng2024bayesoptimalfair} have investigated the Bayes-optimal fair  classifiers with multiple sensitive features, their analysis restricted to the setting of \textit{perfect fairness}, which requires exact equality across groups.
While considered as an ideal target,  perfect fairness is often infeasible on finite data and rarely required in practice \citep{agarwal2018reductions,makhlouf2021machine}. 
\textit{Approximate fairness} acknowledges these limitations and offers a more practical alternative.
It has been widely adopted in research and aligns with practical rules and policy guidelines, such as those from the U.S. Equal Employment Opportunity Commission (EEOC) and the NIST standard \citep{schwartz2022towards}.

Recent work by \citet{chen2024posthoc} and \citet{xian2024unifiedpostprocessing} extended the exploration of Bayes-optimal fair classifiers to approximate fairness settings with multiple sensitive features.
However, these works focus  exclusively on post-processing methods. 
While post-processing approaches are easy to implement, they provide no direct control over the learning objective and may raise legal and interpretability concerns \citep{caton2024fairness}. 
For instance, modifying model outputs post hoc may have legal implications \citep{Barocas2016} and conflict with legal expectations around transparency and explainability \citep{ lum2016statistical,lepri2018fair}.
Moreover, these works are limited to the MD measure and do not address fairness notions like \textit{accuracy parity}.
In contrast, our framework supports both MD \& MR measures, and includes  both in- \& post-processing solutions.
This offers practitioners greater flexibility in choosing \textit{when} to apply fairness interventions and \textit{which} measure to adopt, depending their practical needs. 
In addition, we accommodate fairness notions (including accuracy parity) beyond those addressed in prior work.

Among these, the work most closely related to ours is \citet{zeng2024bayesoptimalfair}, which proposes both in- and post-processing methods for recovering Bayes-optimal fair classifiers. 
While they mentioned possible extension of their framework to multi-class sensitive features, it still focuses on perfect fairness and is hard to address approximate fairness for multiple sensitive features. 
Under perfect fairness, the number of equality constraints is known, allowing their direct use of Neyman–Pearson lemma. 
However, for approximate fairness with multi-sensitive features, the number of active constraints (equalities vs. inequalities) is unknown. 
This presents challenges in both analysis and optimization, thereby limiting the extensibility of their approach.
Our work accommodates  both perfect fairness and approximate fairness (under MD \& MR measures) for multiple sensitive features. 
We also address accuracy parity, beyond those addressed in their work.

\section{Extensions}\label{apxsec:Extensions}

\subsection{Independent Group Fairness \& Intersectional Group Fairness}\label{apxsec:Independent_Intersectional_Extensions}
In real-life situations, a person might be coded with multiple sensitive features.
However, the coexistence of multiple sensitive features (e.g., race and gender simultaneously) gives rise to disparate interpretations of group fairness \citep{Yang2020}.

From one perspective, fairness can be evaluated separately for each sensitive feature, resulting in overlapping subgroups. 
For instance, one can enforce \textit{demographic parity} among subgroups defined by race and simultaneously ensure \textit{demographic parity} among subgroups categorized by gender. 
This type of fairness refers to \textit{independent group fairness} for multiple sensitive features.

Alternatively, another viewpoint entails examining all subgroups defined by the intersections of sensitive features, such as race and gender combined (e.g., white males, white females, males of color, and females of color, etc.). 
    This one refers to \textit{intersectional group fairness} \citep{gohar2023survey}.
    Intersectional fairness is free from the double-counting issue.
    Besides, enforcing intersectional fairness can control independent fairness \citep{Yang2020}, but the converse is not always true \citep{kearns2018preventing}.
    Thus, \textit{intersectional group fairness} is often considered the ideal approach. 
    
    {In real life, intersectional fairness may face the challenge of data sparsity in small subgroups, especially when the number  of sensitive features increases.
    To mitigate this issue, several standard techniques for handling imbalanced data, such as up-sampling, down-sampling, or cost-sensitive learning (or other imbalanced learning methods) can be naturally combined (with our framework).
    For example, during the estimation of group membership probabilities (in the proposed algorithms), up-sampling can help ensure better estimates for small subgroups.}

\subsection{Independent Group Fairness with Multiple Sensitive Features}\label{apxsec:extension_indepent}

As discussed in Section~\ref{sec:Fairness-aware_Learning_Binary_Classification}, multiple sensitive features can be addressed via two approaches: independent group fairness and intersectional group fairness. 
While our main results focus on intersectional fairness, they can be readily extended to the independent fairness setting. 
Below, we illustrate this extension using the MD approximate fairness measure as an example.

In this paper, we consider $K$ sensitive features, where the $k$-th feature takes values 
$m_k \in \mathcal{A}_k$ for $k \in [K]$. Let $M_k = |\mathcal{A}_k|$ be 
the total number of subgroups defined by the $k$-th sensitive feature.
For each sensitive feature $A_k$, we impose a MD fairness constraint:
$$
  \mathrm{MD}_k(f) 
  = 
  \max_{m_k \in [M_k]} 
  \max\left(\mathrm{MD}_{m_k}(f), \mathrm{MD}_{m_k}(1-f)\right)
  \leq \delta,
$$
where $\mathrm{MD}_{m_k}(f)$ is defined analogously to the multi-class sensitive feature case as: $\mathrm{MD}_{m_k}(f)= P(\mathcal{G}(\hat{Y}_{f}) \mid Z=z) - P(\mathcal{G}(\hat{Y}_{f}) \mid Z=z, A_{k}=m_{k})$.
Thus, the \textit{independent group fairness} requirement imposes these 
$K$ separate constraints simultaneously:
$$\mathrm{MD}_k(f)\leq\delta,
  \quad \forall k \in [K].$$
Since Lemma~\ref{lem:MD_FNR_risk} holds for each $\mathrm{MD}_k(f)\leq\delta$, the term $R^\mathrm{MD}_{m_{k}}(f)$ can be derived from Lemma~\ref{lem:MD_FNR_risk} with $m$ replaced by $m_{k}$.
By applying the Lagrangian principle, we can extend Lemma~\ref{lem:Lagrangian_formlation} and Theorem~\ref{th:Fairbayes_form_MD} to the independent fairness setting as follows.

\begin{corollary}
\label{corl:Lagrangian_formlation_indep}
Let $\mathcal{D}$ be any distribution, and consider a cost-sensitive risk 
$R_{cs}(f; c)$ and the MD fairness measure. For any $c \in [0,1]$ and 
$\delta \in [0,1]$, there exist vectors of multipliers 
$\bs{\lambda}_k \in \mathbbm{R}^{M_k}$ (one for each sensitive feature $k$) such that
$$
  \min_{f \in \mathcal{F}} 
    \left\{
      R_{cs}(f; c) 
      : \mathrm{MD}_k(f) \leq \delta,\forall k \in [K]
    \right\}
  \equiv
  \min_{f \in \mathcal{F}} 
    \left(
      R_{cs}(f; c) 
      -
      \sum_{k=1}^{K}\sum_{m_k=1}^{M_k} 
          \lambda_{m_k}   R^\mathrm{MD}_{m_k}(f)
    \right),
$$
where $\lambda_{m_k}$ is the $m_k$-th component of $\bs{\lambda}_k$.
\end{corollary}

\begin{corollary}
\label{corl:Fairbayes_form_MD_indep}
Suppose $c,\delta \in [0,1]$. Then there exists a collection of multiplier 
vectors $\{\bs{\lambda}_k\}_{k=1}^K$ with $\bs{\lambda}_k \in \mathbbm{R}^{M_k}$ such that any Bayes-optimal classifier
$$
  f_{B}^{*}(x)
  \in
  \argmin_{f\in\mathcal{F}}
  \left\{
    R_{cs}(f; c) 
    : 
    \mathrm{MD}_k(f) \leq \delta,\forall k \in [K]
  \right\}
$$
has the threshold form
$$
  f_{B}^{*}(x)
  =
  \mathbbm{1}\left[
    \eta(x) 
    -
    c
    -
    \sum_{k=1}^{K}
    \sum_{y\in\{0,1\}}
    \sum_{m_k=1}^{M_k}
      b_{m_k}^{y}\left(\lambda_{m_k}  - \Lambda_{M_k} a_{m_k}\right)
      \cdot {\gamma_{m_k}^{y}(x)}
    > 
    0
  \right],
$$
where: 
\begin{itemize}[leftmargin=2em, itemsep=5pt, parsep=0pt, topsep=0pt]
    \item[-] $\eta(x) = P(Y=1 \mid X=x)$; 
\item[-]$\gamma_{m_{k}}^{y}(x)=\dfrac{P(E_{y,m_{k}}\mid X=x)}{P(E_{y,m_{k}})}=\dfrac{P(Y=y,A_{k}=m_{k}\mid X=x)}{P(Y=y,A_{k}=m_{k})}$;
\item[-] $\Lambda_{M_k} = \sum_{m_k=1}^{M_k} \lambda_{m_k}$;  
\item[-] The values of constants $a_{m_k}$ and $b_{m_k}^y$ follow from the fairness notions in Table \ref{tab:genernal_fair_measure}.
\end{itemize}

\end{corollary}

The above results mirror the intersectional fairness case from the main text, 
with the key difference being that each sensitive feature $A_k$ has its own set of 
fairness constraints and associated Lagrange multipliers.
Similar results can be derived for the attribute-aware setting as well as for the MR approximate fairness measure using analogous arguments.

\subsection{Composite Fairness Notions}\label{apxsec:extension_composite}
Our framework can also be easily extended to \emph{composite fairness notions} such as \textit{Equalized Odds} \citep{Hardt2016}.
We illustrate this for the MD version of \textit{Equalized Odds}. 

\textit{Equalized Odds} requires that each sensitive group $m \in [M]$ have the same False Positive Rate (\textit{Predictive Equality}) and the same True Positive Rate (\textit{Equal Opportunity}) as the overall population. Concretely, for any classifier outputs prediction $\hat{Y}_{f}$,
\begin{align}\label{eq:apx_EO_def}
P(\hat{Y}_{f} = 1 \mid Y = 1)&=P(\hat{Y}_{f} = 1 \mid S = m, Y = 1),\\\label{eq:apx_PE_def}
P(\hat{Y}_{f} = 1 \mid Y = 0)&=P(\hat{Y}_{f} = 1 \mid S = m, Y =0).
\end{align}
In other words, it requires both \textit{Equal Opportunity} and \textit{Predictive Equality} constraints  hold simultaneously.
Therefore, we can apply the MD measure for Equal Opportunity (Eq. \eqref{eq:apx_EO_def}) and Predictive Equality (Eq. \eqref{eq:apx_PE_def}) simultaneously on the sensitive feature $S$: 
\begin{align}\nohao
\mathrm{MD}^{\mathrm{EO}}(f)&=\max_{m \in [M]}  \max\left(\mathrm{MD}_{m}^{\mathrm{EO}}(f), \mathrm{MD}_{m}^{\mathrm{EO}}(1-f)\right) \leq \delta,\\\nohao
\mathrm{MD}^{\mathrm{PE}}(f)&=\max_{m \in [M]} \max\left(\mathrm{MD}_{m}^{\mathrm{PE}}(f), \mathrm{MD}_{m}^{\mathrm{PE}}(1-f)\right) \leq \delta,
\end{align}
where $\mathrm{MD}_{m}^{\mathrm{EO}}$ and $\mathrm{MD}_{m}^{\mathrm{PE}}$ are defined as in Definition \ref{df:mean_diff} and Table \ref{tab:genernal_fair_measure} (i.e., $Z=Y$ and $z=1$ for EO, and $z=0$ for PE).
Hence, \textit{Equalized Odds} imposes both of the above constraints simultaneously.

Since Lemma~\ref{lem:MD_FNR_risk} holds for each fairness notion, the term $R^{\mathrm{MD,EO}}_{m}(f)$ and $R^{\mathrm{MD,PE}}_{m}(f)$ can be derived by Lemma~\ref{lem:MD_FNR_risk} with selecting the corresponding values of $a_{m}$, $b_{m}^{y}$ and $c_{m}^{y}$.
By applying the Lagrangian principle (as in Lemma~\ref{lem:Lagrangian_formlation} for single fairness constraint), we introduce two sets of Lagrange multipliers—one set for the EO constraint and one set for the PE constraint.  The result is as follows:
\begin{corollary}\label{corl:apx_Lagrangian_formlation}
Let $\mathcal{D}$ be any distribution, and consider the cost-sensitive risk $R_{cs}(f; c)$.  For any $c \in [0,1]$ and $\delta \in [0,1]$, there exist vectors of multipliers 
$\bs{\lambda}^{\mathrm{EO}},\bs{\lambda}^{\mathrm{PE}} \in \mathbbm{R}^{M}$
such that
\begin{align*}
  &\min_{f \in \mathcal{F}} 
    \left\{
      R_{cs}(f; c) 
      : 
      \mathrm{MD}^{\mathrm{EO}}(f) \leq \delta,
      \mathrm{MD}^{\mathrm{PE}}(f) \leq \delta
    \right\}\\
  \equiv&
  \min_{f \in \mathcal{F}} 
    \left(
      R_{cs}(f; c) 
      -
      \sum_{m=1}^{M} \lambda_m^{\mathrm{EO}}  R^{\mathrm{MD,EO}}_{m}(f)
      -
      \sum_{m=1}^{M} \lambda_m^{\mathrm{PE}}  R^{\mathrm{MD,PE}}_{m}(f)
    \right).
\end{align*}
Here, $\lambda_m^{\mathrm{EO}}$ is the $m$-th component of $\bs{\lambda}^{\mathrm{EO}}$, and $\lambda_m^{\mathrm{PE}}$ is the $m$-th component of $\bs{\lambda}^{\mathrm{PE}}$.
\end{corollary}

We can similarly extend Theorem~\ref{th:Fairbayes_form_MD} to derive the form of the \emph{Bayes-optimal} classifier under \textit{Equalized Odds} as follows, with the values of $b_{m}^{y}$ already plugged in:
\begin{corollary}\label{corl:apx_Fairbayes_form_MD}
For any $c\in[0,1]$ and $\delta\in[0,1]$, there exist
$\bs{\lambda}^{\mathrm{EO}},\bs{\lambda}^{\mathrm{PE}} \in \mathbbm{R}^{M}$
such that any Bayes-optimal classifier
$$
  f_{B}^{*}(x)
  \in
  \argmin_{f\in\mathcal{F}}
  \left\{
    R_{cs}(f; c) 
    : 
    \mathrm{MD}^{\mathrm{EO}}(f) \leq \delta,
    \mathrm{MD}^{\mathrm{PE}}(f) \leq \delta
  \right\}
$$
has the following threshold form:
$$
  f_{B}^{*}(x)
  =
  \mathbbm{1}\left[
    \eta(x) 
    - 
    c
    - 
    \hspace{-3pt}\sum_{m=1}^{M}
      \left(\lambda_{m}^{\mathrm{EO}} 
        \hspace{-2pt}- 
        \Lambda_{M}^{\mathrm{EO}} a_{m}^{\mathrm{EO}}\right)
       \cdot{\gamma_{m}^{\mathrm{EO}}(x)}
    - 
    \hspace{-3pt}\sum_{m=1}^{M}
      \left(\lambda_{m}^{\mathrm{PE}}
        \hspace{-2pt}-
        \Lambda_{M}^{\mathrm{PE}} a_{m}^{\mathrm{PE}}\right)
       \frac{\gamma_{m}^{\mathrm{PE}}(x)}{P(E_{0,m})}
    >
    0
  \right],
$$
where:
\begin{itemize}[leftmargin=2em, itemsep=8pt, parsep=0pt, topsep=2pt]
    \item[-] $\eta(x)=P(Y=1\mid X=x)$;
    \item[-] $\lambda_{m}^{\mathrm{EO}}$ is the $m$-th component of $\bs{\lambda}^{\mathrm{EO}}$, and $\lambda_{m}^{\mathrm{PE}}$ is the $m$-th component of $\bs{\lambda}^{\mathrm{PE}}$;
    \item[-] $\Lambda_{M}^{\mathrm{EO}} = \sum_{m=1}^{M}\lambda_{m}^{\mathrm{EO}}$ and $\Lambda_{M}^{\mathrm{PE}} = \sum_{m=1}^{M}\lambda_{m}^{\mathrm{PE}}$;
    \item[-] $\gamma_{m}^{\mathrm{EO}}(x) = \frac{P(E_{1,m}\mid X=x)}{P(E_{1,m})}=\frac{P(Y=1, S=m\mid X=x)}{P(Y=1, S=m)}$ and $\gamma_{m}^{\mathrm{PE}}(x) = \frac{P(E_{0,m}\mid X=x)}{P(E_{0,m})}=\frac{P(Y=0, S=m\mid X=x)}{P(Y=0, S=m)}$;
    \item[-] $a_{m}^{\mathrm{EO}} = P_{S\mid Y =1}(m)$ and $a_{m}^{\mathrm{PE}} = P_{S\mid Y =0}(m)$ 
  (see Table~\ref{tab:genernal_fair_measure} for details).
\end{itemize}
\end{corollary}

In a similar manner, one can adapt these arguments to: 1) MR approximate fairness measures; 2) The {attribute-aware} setting; and 3) Other composite fairness notions which combine two or more fairness notions from Table~\ref{tab:genernal_fair_measure} (similar to \textit{Equalized Odds}, the results are extended by incorporating additional instance-dependent threshold correction terms into the decision boundary, each corresponding to a specific fairness notion).

\section{Proofs}\label{apxsec:Proofs}
\subsection{Proof of Lemma \ref{lem:MD_FNR_risk}}

\begin{proof}
By definition,
$$\mathrm{MD}(f) = \max_{m\in [M]} \max\left(\mathrm{MD}_{m}(f), \mathrm{MD}_{m}(1-f)\right).$$
Hence, $\mathrm{MD}(f)\le\delta$ 
is equivalent to 
$\mathrm{MD}_m(f)\le\delta$ \emph{and} $\mathrm{MD}_m(1-f)\le\delta$ for all $m\in[M]$.

From Table~\ref{tab:genernal_fair_measure}, one verifies that 
$$
  \mathrm{MD}_m(1-f) +\mathrm{MD}_m(f) = 0
  \quad
  \text{for each fairness notion}.
$$
Thus, 
$\mathrm{MD}_m(f)\le\delta$ and $\mathrm{MD}_m(1-f)\le\delta$ together imply
$\mathrm{MD}_m(f)\in[-\delta,\delta]$.  
Hence, $\mathrm{MD}(f)\le\delta$ if and only if $\mathrm{MD}_m(f)\in[-\delta,\delta]$ for every $m$.
We define $R^\mathrm{MD}_{m}(f) = \mathrm{MD}_m(f)$, then $
  \mathrm{MD}(f)\le\delta
  \Longleftrightarrow
  R^\mathrm{MD}_{m}(f)\in[-\delta,\delta]$ for $\forall m\in[M].
$

Next, we express  $\mathrm{MD}_m(f)$ in terms of $\hat{Y}_f$ and $Y$. 
One can easily check that for all four fairness notions,
\begin{align}\label{eq:apx_MD_abc_repeated}
  \mathrm{MD}_m(f)=
  \sum_{y \in\{0,1\}} \Bigg[ \tilde{a}^y P\left(\hat{Y}_f=1 \mid Y=y\right)-\tilde{b}_m^y P\left(\hat{Y}_f=1 \mid Y=y, S=m\right)+\widetilde{c}_m^y\Bigg],
\end{align}
where $\tilde{a}^y$, $\tilde{b}_m^y$ and $\widetilde{c}_m^y$ are specified in Table \ref{tab:apx_genernal_measure}.

\begin{table*}[t]

\begin{center}
\resizebox{\textwidth}{!}{%
\begin{tabular}{ccccc}
\toprule
Notion & $\tilde{a}^y$ & $\tilde{b}_m^y$ & $\widetilde{c}_m^y$ (MD) & $\widetilde{c}_m^y$ (MR) \\
\midrule
DP    & $P(Y=y)$ & $P_{Y\mid S =m}(y)$ & $0$ & $0$ \\
EO    & $y$ & $y$ & $0$ & $0$ \\
PE    & $1 - y$ & $1 - y$ & $0$ & $0$ \\
{AP}   & {$(1-2y)P(Y=y)$} & {$(1 - 2y) \cdot P_{Y\mid S =m}(y)$} & $(1-y)p^{+}-yP_{Y\mid S =m}(1)$ & $\delta (1-y) p^{+}-yP_{Y\mid S =m}(1)$ \\
\bottomrule
\end{tabular}
}
\end{center}
\caption{The values of $\tilde{a}^y$, $\tilde{b}_m^y$ and $\widetilde{c}_m^y$ for standard group fairness notions.}
\label{tab:apx_genernal_measure}
\end{table*}

Recall that $E_{y,m}=\{Y=y, S=m\}$. Using the law of total probability: \begin{align}\nohao
P\left(\hat{Y}_f=1 \mid Y=y\right)&=\sum_{m'=1}^{M}P\left(\hat{Y}_f=1\mid  Y=y, S=m'\right)\cdot P(S=m'\mid Y=y)\\ \label{eq:apx_lem3.2_eq1}&=\sum_{m'=1}^{M}P\left(\hat{Y}_f=1\mid E_{y,m'}\right)\cdot P(S=m'\mid Y=y).    
\end{align} 
Since $R^\mathrm{MD}_{m}(f)=\mathrm{MD}_m(f)$, plugging \eqref{eq:apx_lem3.2_eq1} into \eqref{eq:apx_MD_abc_repeated} yields:
\begin{align}\nohao
 R^\mathrm{MD}_{m}(f)=
  \sum_{y \in\{0,1\}} &\left\{\Big[  \sum_{m'=1}^{M} \tilde{a}^y P\left(\hat{Y}_f=1\mid E_{y,m'}\right)\cdot P(S=m'\mid Y=y)\Big]\right.\\\label{eq:apx_lem3.2_eq2}&  -\tilde{b}_m^y P\left(\hat{Y}_f=1 \mid E_{y,m}\right)+\widetilde{c}_m^y \Bigg\}. 
\end{align}

\noindent\textbf{For DP and AP:}  \eqref{eq:apx_lem3.2_eq2} can further reduce to 
\begin{align}\nohao 
 R^\mathrm{MD}_{m}(f)&=
  \sum_{y \in\{0,1\}} \left\{\Big[  \sum_{m'=1}^{M} \tilde{a}^y \cdot \dfrac{P_{Y\mid S =m'}(y)P(S=m')}{P(Y=y)} \cdot P\left(\hat{Y}_f=1\mid E_{y,m'}\right)\Big]\right.\\\nohao
  &\qquad\qquad -\tilde{b}_m^y P\left(\hat{Y}_f=1 \mid E_{y,m}\right)+\widetilde{c}_m^y \Bigg\}\\\nohao
  &=\sum_{y \in\{0,1\}} \left\{\Big[  \sum_{m'=1}^{M} a_{m'} b_{m'}^{y} \cdot P\left(\hat{Y}_f=1\mid E_{y,m'}\right)\Big]- {b}_m^y P\left(\hat{Y}_f=1 \mid E_{y,m}\right)+ {c}_m^y \right\},
\end{align}
where
\begin{itemize}[leftmargin=2em]
    \item \textbf{DP:} $a_m =\tilde{a}^y\dfrac{P(S=m)}{P(Y=y)}=P_{S}(m)$, ${b}_m^y =\tilde{b}_m^y$ and ${c}_m^y =\widetilde{c}_m^y$ for $\forall m\in[M]$, since  $\tilde{b}_m^y=P_{Y\mid S =m}(y)$ for DP.
    \item \textbf{AP:} $a_m =\dfrac{\tilde{a}^y\cdot P(S=m)}{(1-2y)P(Y=y)}=P_{S}(m)$, ${b}_m^y =\tilde{b}_m^y$ and ${c}_m^y =\widetilde{c}_m^y$ for $\forall m\in[M]$, since $\tilde{b}_m^y=(1-2y)\cdot P_{Y\mid S =m}(y)$ for AP. 
\end{itemize}

\noindent\textbf{For EO and PE:} Note that $\tilde{a}^y=\tilde{b}_m^y$ for any $m$. Thus, \eqref{eq:apx_lem3.2_eq2} can further reduce to 
\begin{align}\nohao
  R^\mathrm{MD}_{m}(f)&=
  \sum_{y \in\{0,1\}} \left\{\Big[  \sum_{m'=1}^{M} \tilde{b}_{m'}^y P(S=m'\mid Y=y) P\left(\hat{Y}_f=1\mid E_{y,m'}\right)\Big]\right.\\\nohao
  &\qquad\qquad-\tilde{b}_m^y P\left(\hat{Y}_f=1 \mid E_{y,m}\right)+\widetilde{c}_m^y \Bigg\}\\\nohao
  &=\sum_{y \in\{0,1\}} \left\{\Big[  \sum_{m'=1}^{M} a_{m'} b_{m'}^{y} \cdot P\left(\hat{Y}_f=1\mid E_{y,m'}\right)\Big]- {b}_m^y P\left(\hat{Y}_f=1 \mid E_{y,m}\right)+ {c}_m^y \right\},
\end{align}
where 
\begin{itemize}[leftmargin=2em]
\item \textbf{EO:} $a_{m}=P_{S\mid Y=1}(m)$,  ${b}_m^y=\tilde{b}_m^y$ and ${c}_m^y:=\widetilde{c}_m^y$ for $\forall m\in[M]$ (\textit{Note}: Since $\tilde{b}_m^0\equiv0$ for EO, it follows that $\tilde{b}_{m'}^0 P(S=m'\mid Y=0)=\tilde{b}_{m'}^0 P(S=m'\mid Y=1)$).

\item \textbf{PE:} $a_{m}=P_{S\mid Y=0}(m)$, ${b}_m^y=\tilde{b}_m^y$ and ${c}_m^y:=\widetilde{c}_m^y$ for $\forall m\in[M]$ (\textit{Note}: Since $\tilde{b}_m^1\equiv0$ for PE, it follows that $\tilde{b}_{m'}^1 P(S=m'\mid Y=1)=\tilde{b}_{m'}^1 P(S=m'\mid Y=0)$).

\end{itemize}

Therefore, for all four fairness notions, we have 
$$
  \mathrm{MD}(f)\le\delta
  \Longleftrightarrow
  \mathrm{MD}_m(f)\in[-\delta, \delta]\quad\forall m\in[M]
  \Longleftrightarrow
  R^\mathrm{MD}_{m}(f)\in[-\delta,\delta]\quad\forall m\in[M],
$$
with $$R^\mathrm{MD}_{m}(f)=\sum_{y \in\{0,1\}} \left\{\Big[  \sum_{m'=1}^{M} a_{m'} b_{m'}^{y} \cdot P\left(\hat{Y}_f=1\mid E_{y,m'}\right)\Big]- {b}_m^y P\left(\hat{Y}_f=1 \mid E_{y,m}\right)+ {c}_m^y \right\},$$
where the values of $a_{m}$, $b_{m}^{y}$, and $c_{m}^{y}$ are as shown in Table \ref{tab:genernal_fair_measure}.
This completes the proof.
\end{proof}

\subsection{Proof of Lemma \ref{lem:MR_FNR_risk}}
\begin{proof}
By definition,
$$\mathrm{MR}(f)=\min_{m\in [M]} \min\left(\mathrm{MR}_{m}(f), \mathrm{MR}_{m}(1-f)\right).$$
Hence, $\mathrm{MR}(f)\geq\delta$ 
is equivalent to 
$\mathrm{MR}_m(f)\geq\delta$ \emph{and} $\mathrm{MR}_m(1-f)\geq\delta$ for all $m\in[M]$.

From Table~\ref{tab:genernal_fair_measure}, one verifies that for each fairness notion,
$$\mathrm{MR}_m(f)\geq\delta \Longleftrightarrow \delta\cdot{P(\mathcal{G}(\hat{Y}_{f}) \mid Z=z)}-{P(\mathcal{G}(\hat{Y}_{f}) \mid Z=z, S = m)}\leq 0,
$$
$$\mathrm{MR}_m(1-f)\geq\delta \Longleftrightarrow \delta\cdot{P(\mathcal{G}(\hat{Y}_{f}) \mid Z=z)}-{P(\mathcal{G}(\hat{Y}_{f}) \mid Z=z, S = m)}\geq \delta-1.
$$
Define $R^\mathrm{MR}_{m}(f)=\delta\cdot{P(\mathcal{G}(\hat{Y}_{f}) \mid Z=z)}-{P(\mathcal{G}(\hat{Y}_{f}) \mid Z=z, S = m)}$.
Then, 
$\mathrm{MR}_m(f)\geq\delta$ and $\mathrm{MR}_m(1-f)\geq\delta$ together imply
$R^\mathrm{MR}_{m}(f)\in[\delta-1,0]$.  
Hence, $\mathrm{MR}(f)\geq\delta$ if and only if $R^\mathrm{MR}_{m}(f)\in[\delta-1,0]$ for $\forall m\in[M]$.

Next, we express $\mathrm{MR}_m(f)$ in terms of $\hat{Y}_f$ and $Y$ analogously to the MD case. 
One can easily check that for all four fairness notions,
\begin{align}\label{eq:apx_MR_abc_repeated}
  R^\mathrm{MR}_{m}(f)=
  \sum_{y \in\{0,1\}} \Bigg[ \delta \cdot \tilde{a}^y  P\left(\hat{Y}_f=1 \mid Y=y\right)-\tilde{b}_m^y P\left(\hat{Y}_f=1 \mid Y=y, S=m\right)+\widetilde{c}_m^y\Bigg],
\end{align}
where $\tilde{a}^y$, $\tilde{b}_m^y$ and $\widetilde{c}_m^y$ are specified in Table \ref{tab:apx_genernal_measure}.

Then, plugging \eqref{eq:apx_lem3.2_eq1} into \eqref{eq:apx_MR_abc_repeated} yields:
\begin{align}\nohao
 R^\mathrm{MR}_{m}(f)=
  \sum_{y \in\{0,1\}} &\left\{\delta \cdot\Big[  \sum_{m'=1}^{M}  \tilde{a}^y P\left(\hat{Y}_f=1\mid E_{y,m'}\right)\cdot P(S=m'\mid Y=y)\Big]\right.\\\label{eq:apx_lem3.4_eq2}
  &-\tilde{b}_m^y P\left(\hat{Y}_f=1 \mid E_{y,m}\right)+\widetilde{c}_m^y \Bigg\}, 
\end{align}
where $E_{y,m}=\{Y=y, S=m\}$.

\noindent\textbf{For DP and AP:}  \eqref{eq:apx_lem3.4_eq2} can further reduce to 
\begin{align}\nohao 
R^\mathrm{MR}_{m}(f)&=
  \sum_{y \in\{0,1\}} \left\{\delta \cdot\Big[\sum_{m'=1}^{M} \tilde{a}^y \cdot \dfrac{P_{Y\mid S =m'}(y)P(S=m')}{P(Y=y)} \cdot P\left(\hat{Y}_f=1\mid E_{y,m'}\right)\Big]\right.\\\nohao& \qquad\qquad-\tilde{b}_m^y P\left(\hat{Y}_f=1 \mid E_{y,m}\right)+\widetilde{c}_m^y \Bigg\}\\\nohao
  &=\sum_{y \in\{0,1\}} \left\{\delta \cdot\Big[  \sum_{m'=1}^{M} a_{m'} b_{m'}^{y} \cdot P\left(\hat{Y}_f=1\mid E_{y,m'}\right)\Big]- {b}_m^y P\left(\hat{Y}_f=1 \mid E_{y,m}\right)+ {c}_m^y \right\},
\end{align}
where
\begin{itemize}[leftmargin=2em]
    \item \textbf{DP:} $a_m =\tilde{a}^y\dfrac{P(S=m)}{P(Y=y)}=P_{S}(m)$, ${b}_m^y =\tilde{b}_m^y$ and ${c}_m^y =\widetilde{c}_m^y$ for $\forall m\in[M]$, since  $\tilde{b}_m^y=P_{Y\mid S =m}(y)$ for DP.
    \item \textbf{AP:} $a_m =\dfrac{\tilde{a}^y\cdot P(S=m)}{(1-2y)P(Y=y)}=P_{S}(m)$, ${b}_m^y =\tilde{b}_m^y$ and ${c}_m^y =\widetilde{c}_m^y$ for $\forall m\in[M]$, since $\tilde{b}_m^y=(1-2y)\cdot P_{Y\mid S =m}(y)$ for AP. 
\end{itemize}

\noindent\textbf{For EO and PE:} Note that $\tilde{a}^y=\tilde{b}_m^y$ for any $m$. Thus, \eqref{eq:apx_lem3.4_eq2} can further reduce to 
\begin{align}\nohao
R^\mathrm{MR}_{m}(f)&=
  \sum_{y \in\{0,1\}} \left\{\delta \cdot\Big[  \sum_{m'=1}^{M} \tilde{b}_{m'}^y P(S=m'\mid Y=y) P\left(\hat{Y}_f=1\mid E_{y,m'}\right)\Big]\right.\\\nohao
  &\qquad\qquad-\tilde{b}_m^y P\left(\hat{Y}_f=1 \mid E_{y,m}\right)+\widetilde{c}_m^y \Bigg\}\\\nohao
  &=\sum_{y \in\{0,1\}} \left\{\delta \cdot\Big[  \sum_{m'=1}^{M} a_{m'} b_{m'}^{y} \cdot P\left(\hat{Y}_f=1\mid E_{y,m'}\right)\Big]- {b}_m^y P\left(\hat{Y}_f=1 \mid E_{y,m}\right)+ {c}_m^y \right\},
\end{align}
where 
\begin{itemize}[leftmargin=2em]
\item \textbf{EO:} $a_{m}=P_{S\mid Y=1}(m)$,  ${b}_m^y=\tilde{b}_m^y$ and ${c}_m^y:=\widetilde{c}_m^y$ for $\forall m\in[M]$ (\textit{Note}: Since $\tilde{b}_m^0\equiv0$ for EO, it follows that $\tilde{b}_{m'}^0 P(S=m'\mid Y=0)=\tilde{b}_{m'}^0 P(S=m'\mid Y=1)$).

\item \textbf{PE:} $a_{m}=P_{S\mid Y=0}(m)$, ${b}_m^y=\tilde{b}_m^y$ and ${c}_m^y:=\widetilde{c}_m^y$ for $\forall m\in[M]$ (\textit{Note}: Since $\tilde{b}_m^1\equiv0$ for PE, it follows that $\tilde{b}_{m'}^1 P(S=m'\mid Y=1)=\tilde{b}_{m'}^1 P(S=m'\mid Y=0)$).

\end{itemize}

Therefore, for all four fairness notions, we have  
$$
  \mathrm{MR}(f)\geq\delta
  \Longleftrightarrow
  R^\mathrm{MR}_{m}(f)\in[\delta-1,0]\quad\forall m\in[M].
$$
with $$R^\mathrm{MR}_{m}(f)=\sum_{y \in\{0,1\}} \left\{\delta \cdot\Big[  \sum_{m'=1}^{M} a_{m'} b_{m'}^{y} \cdot P\left(\hat{Y}_f=1\mid E_{y,m'}\right)\Big]- {b}_m^y P\left(\hat{Y}_f=1 \mid E_{y,m}\right)+ {c}_m^y \right\},$$
where the values of $a_{m}$, $b_{m}^{y}$, and $c_{m}^{y}$ are as shown in Table \ref{tab:genernal_fair_measure}.
This completes the proof.
\end{proof}

\subsection{Proof of Lemma \ref{lem:Lagrangian_formlation}}

\begin{proof}
We first address the {MD} measure, followed by analogous steps for the {MR} measure.

\noindent
\textbf{Case 1: MD Measure.}

The MD-fairness constraint (see \eqref{eq:mean_diff_constr}) requires:
$$
  \mathrm{MD}(f)
  =
  \max_{m\in [M]} 
    \max\left(\mathrm{MD}_{m}(f),  \mathrm{MD}_{m}(1-f)\right)
  \leq\delta.
$$
By Lemma~\ref{lem:MD_FNR_risk}, this is equivalent to enforcing 
$$
  R^\mathrm{MD}_{m}(f)\in[-\delta, \delta]
  \quad
  \text{for all } m \in [M],
$$
where $R^\mathrm{MD}_{m}(f)$ is as defined in Lemma~\ref{lem:MD_FNR_risk}.  
Thus, the optimization
$$
  \min_{f \in \mathcal{F}}
    \left\{
      R_{cs}(f; c) : \mathrm{MD}(f) \leq \delta
    \right\}
  \equiv
  \min_{f \in \mathcal{F}}
    \left\{
      R_{cs}(f; c) 
      :
      -\delta \leq R^\mathrm{MD}_{m}(f) \leq \delta 
      \quad\forall m\in[M]
    \right\}.
$$

{By Lemma~\ref{apx:suppt_lem_linear_program}, the right-hand side of the above equation is a linear program in $f$.
 Strong duality for linear programs\footnote{This implicitly assumes feasibility of the primal problem, i.e. for a pre-specified $\delta$, we need that there exists at least one (randomized) classifier $f$ with symmetrized fairness level at {most} $\delta$. 
Notably, for {DP}, {EO}, and {PE}, feasibility is always guaranteed by the trivial {constant} classifier (which predicts the same label for all inputs).} 
tells us that
\begin{align}\nohao
&\min_{f \in \mathcal{F}}
    \left\{
      R_{cs}(f; c) 
      :
      -\delta \leq R^\mathrm{MD}_{m}(f) \leq \delta 
      \quad\forall m\in[M]
    \right\}\\\nohao
    = &\max_{\lambda^{1}_{m}, \lambda^{2}_{m}\geq 0}\min_{f \in \mathcal{F}} \left\{R_{cs}(f; c) + \sum_{m=1}^{M} \Big[\lambda^{1}_{m} \cdot \left(R^\mathrm{MD}_{m}(f)-\delta\right)- \lambda^{2}_{m}\left(R^\mathrm{MD}_{m}(f)+\delta\right)\Big]\right\}\\\nohao
    =&\min_{f \in \mathcal{F}} \left\{R_{cs}(f; c) + \sum_{m=1}^{M} \Big[\lambda^{1*}_{m} \cdot \left(R^\mathrm{MD}_{m}(f)-\delta\right)- \lambda^{2*}_{m}\left(R^\mathrm{MD}_{m}(f)+\delta\right)\Big]\right\}, 
\end{align}
where $$\lambda^{1*}_{m},\lambda^{2*}_{m}=\argmax_{\lambda^{1}_{m}, \lambda^{2}_{m}\geq 0}\min_{f \in \mathcal{F}} \left\{R_{cs}(f; c) + \sum_{m=1}^{M} \Big[\lambda^{1}_{m} \cdot \left(R^\mathrm{MD}_{m}(f)-\delta\right)- \lambda^{2}_{m}\left(R^\mathrm{MD}_{m}(f)+\delta\right)\Big]\right\}.$$

Therefore, when converting the hard fairness constraint to a soft constraint on the Lagrangian, one will thus obtain two sets of non-negative Lagrange (KKT) multipliers $\lambda^{1*}_{m}\geq 0$ and $\lambda^{2*}_{m}\geq 0$ for each group $m$, so that}
\begin{align}\nohao
&\min_{f \in \mathcal{F}} \left\{R_{cs}(f; c) : \mathrm{MD}(f) \leq \delta\right\} \\\nohao=&\min_{f \in \mathcal{F}} \left\{R_{cs}(f; c) + \sum_{m=1}^{M} \Big[\lambda^{1*}_{m} \cdot \left(R^\mathrm{MD}_{m}(f)-\delta\right)- \lambda^{2*}_{m}\left(R^\mathrm{MD}_{m}(f)+\delta\right)\Big]\right\}\\\nohao
=&\min_{f \in \mathcal{F}} \left\{R_{cs}(f; c) - \sum_{m=1}^{M} 
  (\lambda_m^{2*} -\lambda_m^{1*}) R_m^{\mathrm{MD}}(f)
  - \sum_{m=1}^{M}(\lambda_m^{1*} + \lambda_m^{2*}) \delta
\right\}
\\\nohao
=&\min_{f \in \mathcal{F}} \left\{R_{cs}(f; c) - \sum_{m=1}^{M} 
  (\lambda_m^{2*} -\lambda_m^{1*}) R_m^{\mathrm{MD}}(f)
\right\}.
\end{align}
The last equation holds because 
$\sum_{m=1}^M (\lambda_m^{1*} + \lambda_m^{2*}) \delta
$ does not depend on \( f \), and thus, it does not affect the minimization. 
Defining 
$\lambda_m := \lambda_m^{2*} - \lambda_m^{1*}
$ proves the MD case.

\noindent
\textbf{Case 2: MR Measure.}

The MR-fairness constraint (see \eqref{eq:mean_ratio_constr}) requires:
$$
\mathrm{MR}(f)=\min_{m\in [M]} \min\left(\mathrm{MR}_{m}(f), \mathrm{MR}_{m}(1-f)\right) \geq \delta.
$$
By Lemma~\ref{lem:MR_FNR_risk}, this is equivalent to enforcing 
$$
 R^\mathrm{MR}_{m}(f)\in[\delta-1,0]
  \quad
  \text{for all } m \in [M],
$$
where $R^\mathrm{MR}_{m}(f)$ is defined in Lemma~\ref{lem:MR_FNR_risk}.  
Hence, the optimization
$$
  \min_{f \in \mathcal{F}}
    \left\{
      R_{cs}(f; c) : \mathrm{MR}(f) \geq \delta
    \right\}
  \equiv
  \min_{f \in \mathcal{F}}
    \left\{
      R_{cs}(f; c) 
      :
      \delta-1 \leq R^\mathrm{MR}_{m}(f) \leq 0 
      \quad\forall m\in[M]
    \right\}.
$$
By Lemma~\ref{apx:suppt_lem_linear_program}, the right-hand side of the above equation is a linear program in $f$. 
Similar to the MD case, with strong duality, when converting the hard fairness constraint to a soft constraint on the Lagrangian, one will thus obtain two sets of non-negative Lagrange (KKT) multipliers $\lambda^{1*}_{m}\geq 0$ and $\lambda^{2*}_{m}\geq 0$ for each group $m$, so that
\begin{align}\nohao
&\min_{f \in \mathcal{F}} \left\{R_{cs}(f; c) : \mathrm{MR}(f) \geq \delta\right\}\\\nohao =&\min_{f \in \mathcal{F}} \left\{R_{cs}(f; c) + \sum_{m=1}^{M} \Big[\lambda^{1*}_{m} \cdot R^\mathrm{MR}_{m}(f)- \lambda^{2*}_{m} \left(R^\mathrm{MR}_{m}(f)+1-\delta\right) \Big]\right\}\\\nohao
=&\min_{f \in \mathcal{F}} \left\{R_{cs}(f; c) - \sum_{m=1}^{M} 
  (\lambda_m^{2*} -\lambda_m^{1*}) R^\mathrm{MR}_{m}(f)
  + \sum_{m=1}^{M} \lambda_m^{2*} (\delta-1)
\right\}
\\\nohao
=&\min_{f \in \mathcal{F}} \left\{R_{cs}(f; c) - \sum_{m=1}^{M} 
  (\lambda_m^{2*} -\lambda_m^{1*}) R_m^{\mathrm{MR}}(f)
\right\}.
\end{align}
The last equation holds because 
$\sum_{m=1}^{M} \lambda_m^{2*} (\delta-1)$ does not depend on \( f \). 
Defining 
$\lambda_m := \lambda_m^{2*} - \lambda_m^{1*}
$ proves the MR case.

\noindent
Therefore, the lemma holds for both MD and MR measures.
\end{proof}

\subsection{Helper Lemmas}
We first introduce several technical lemmas that are useful for proving the theoretical results presented in Theorems \ref{th:Fairbayes_form_MD} and \ref{th:Fairbayes_form_MR}.

\begin{lemma}\label{apx:suppt_lem_1}
Pick any randomized classifier $f$. 
Denote $\hat{Y}_{f}$ as the prediction of $f$.
Then, for any $m\in[M]$, we have
\begin{equation}
P(\hat{Y}_{f}=1\mid E_{y,m})={E}_{X}\left[{\gamma^{y}_{m}(X)}\cdot f(X)\right],
\end{equation}
where $\gamma^{y}_{m}(X)=\dfrac{P(E_{y,m}\mid X=x)}{P(E_{y,m})}$ and $E_{y,m} = \{Y=y, S=m\}$.
\end{lemma}
\begin{proof}
We first demonstrate $P(\hat{Y}_{f}=1\mid E_{y,m})={E}_{X\mid Y=y, S=m}\left[f(X)\right]$ as follows:
\begin{align}\nohao
\mathbb{E}_{X \mid Y=y, S=m}\left[f(X)\right]
&=
\int_{\mathcal{X}} f(x)  p\left(x \mid Y=y, S=m\right) dx
\\\nohao
&=
\int_{\mathcal{X}} P(\hat{Y}_f = 1 \mid X=x)  p\left(x \mid Y=y, S=m\right) dx
\\\nohao
&\overset{(a)}{=}
\int_{\mathcal{X}} P(\hat{Y}_f=1  \mid  X=x, Y=y, S=m)  p\left(x \mid Y=y, S=m\right) dx
\\\label{eq:apx_suppt_lem_1}
&=
P(\hat{Y}_f = 1  \mid  Y=y, S=m)=P(\hat{Y}_{f}=1\mid E_{y,m}).
\end{align}

Here, $p\left(x \mid Y=y, S=m\right)$ represents the conditional probability density of $X$ given $Y=y$ and $S=m$.
Equality $(a)$ follows from the fact that the distribution of $\hat{Y}_{f}$ is fully determined given $X$.

Next, we check the equivalence:
\begin{align}\nohao
{E}_{X\mid Y=y, S=m}\left[f(X)\right]
&\overset{(b)}{=}{E}_{X}\left[f(X)\dfrac{P(Y=y, S=m\mid X)}{P(Y=y, S=m)}\right]
\\\label{eq:apx_suppt_lem_2}
&{=}{E}_{X}\left[{\gamma^{y}_{m}(X)}\cdot f(X)\right].
\end{align}
The equality (b) holds because the conditional distribution $P(X \mid Y=y, S = m)$ can be expressed as: $P(X \mid Y=y, S = m)=\dfrac{P(X)\cdot P(Y=y, S=m\mid X)}{P(Y=y, S = m)}$.

Finally, combining \eqref{eq:apx_suppt_lem_1} and \eqref{eq:apx_suppt_lem_2} yields the desired result. This completes the proof.
\end{proof}

For completeness, we restate Lemma 9 from \citet{menon2018} as follows.
\begin{lemma}\label{apx:suppt_lem_menon2018}
Pick any randomized classifier $f$. Then, for any cost parameter $c \in [0, 1]$,
$$R_{cs}(f;c) = (1 - c) \cdot {P}(Y = 1) + {E}_{X}\left[(c - \eta(X)) \cdot f(X)\right],$$
where $\eta(x) = {P}(Y = 1 \mid X = x)$.  
\end{lemma}

Lemma \ref{apx:suppt_lem_linear_program} follows directly from Lemma \ref{apx:suppt_lem_menon2018}, as shown below.

\begin{lemma}\label{apx:suppt_lem_linear_program}
Let $c \in [0,1]$ and $\delta \in [0,1]$. Then, the following optimization problems
\begin{align}\nohao
&\min_{f \in \mathcal{F}}
    \left\{
      R_{cs}(f; c) 
      :
      -\delta \leq R^\mathrm{MD}_{m}(f) \leq \delta 
      \quad\forall m\in[M]
    \right\},    \\\nohao
     &\min_{f \in \mathcal{F}}
    \left\{
      R_{cs}(f; c) 
      :
      \delta-1 \leq R^\mathrm{MR}_{m}(f) \leq 0 
      \quad\forall m\in[M]
    \right\}
\end{align}
can both be expressed as linear programs in $f$.  
Here, $\mathcal{F}$ is the set of all measurable classifiers $f: \mathcal{X}\to[0,1]$ (or $\{0,1\}$), viewed as a randomized (or deterministic) predictor.
\end{lemma}
\begin{proof}
According to Lemma \ref{apx:suppt_lem_menon2018}, the cost-sensitive risk $R_{cs}$ is linear in the randomized classier $f$ as follows: 
\begin{align}\nohao
R_{cs}(f;c) &= (1 - c) \cdot p^{+} + {E}_{X}\left[(c - \eta(X)) \cdot f(X)\right]\\\nohao
&=(1 - c) \cdot p^{+} +\int_{\mathcal{X}} \bigl(c - \eta(x)\bigr)f(x)p_X(x)
  dx.
\end{align}
where $p^{+}={P}(Y = 1)$, $p_X(x)$ is the probability density function of $X$, and $\eta(x) = {P}(Y = 1 \mid X = x)$.

Furthermore, by Lemmas \ref{lem:MD_FNR_risk} and \ref{lem:MR_FNR_risk}, both $R^\mathrm{MD}_{m}(f)$ and $R^\mathrm{MR}_{m}(f)$ are linear transformations of $P(\hat{Y}_{f}=1\mid E_{y,m})$. 
Additionally, by Lemma \ref{apx:suppt_lem_1},  $P(\hat{Y}_{f}=1\mid E_{y,m})$ itself is a linear transformation of $f$. 
Therefore, both $R^\mathrm{MD}_{m}(f)$ and $R^\mathrm{MR}_{m}(f)$ are linear functions of $f$.
Specifically, as shown in \eqref{eq:apx_th_bayes_fair_MD2}, for MD,
\begin{align}\nohao
R^\mathrm{MD}_{m}(f)
&=
  \sum_{y\in\{0,1\}} c_m^y
+
  \sum_{y\in\{0,1\}} {E}_X \left[
    -\left( b_m^y {\gamma_{m}^y(X)} -
    \sum_{m'=1}^M a_{m'}b_{m'}^y {\gamma_{m'}^y(X)}\right)\cdot f(X)
  \right]
  \\\nohao
  &=\sum_{y\in\{0,1\}} c_m^y
+
  \sum_{y \in \{0,1\}}
\int_{\mathcal{X}}
  t_{m}^{y}(x)f(x)p_X(x)dx.
\end{align}
where $t_{m}^{y}(x)=
    \sum_{m'=1}^M
      a_{m'}b_{m'}^y\gamma_{m'}^y(x)-b_m^y\gamma_m^y(x)$.
Similarly, as shown in \eqref{eq:apx_th_bayes_fair_MR2}, for MR,
\begin{align}\nohao
R^\mathrm{MR}_m(f) 
&=
  \sum_{y\in\{0,1\}} c_m^y
+
  \sum_{y\in\{0,1\}} {E}_X \left[
    -\left( b_m^y {\gamma_{m}^y(X)} -
    \delta\sum_{m'=1}^M a_{m'}b_{m'}^y {\gamma_{m'}^y(X)}\right)\cdot f(X)
  \right]\\\nohao
&=\sum_{y\in\{0,1\}} c_m^y
+
  \sum_{y \in \{0,1\}}
\int_{\mathcal{X}}
  t_{m}^{y}(x)f(x)p_X(x)dx.
\end{align}
where $t_{m}^{y}(x)=
    \delta\sum_{m'=1}^M
      a_{m'}b_{m'}^y\gamma_{m'}^y(x)-b_m^y\gamma_m^y(x)$.

Now, for $\forall x \in \mathcal{X}$, let $u(x) :=\bigl(c - \eta(x)\bigr)\cdot p_X(x)$, $v_{m}^{y}(x) :=t_{m}^{y}(x) \cdot p_X(x)$, and $W_{m}:=\sum_{y\in\{0,1\}} c_m^y$.
The optimization problems are
\begin{align}\nohao
&\text{For MD:} \quad \min_{f \in \mathcal{F}}
    \left\{
      \int_{\mathcal{X}} u(x)f(x)dx
      :
      -\delta \leq W_{m}+ \sum_{y \in \{0,1\}}
\int_{\mathcal{X}}
  v_{m}^{y}(x)f(x)dx \leq \delta 
      \quad\forall m\in[M]
    \right\},    \\\nohao
     & \text{For MR:} \quad\min_{f \in \mathcal{F}}
    \left\{
       \int_{\mathcal{X}} u(x)f(x)dx 
      :
      \delta-1 \leq W_{m}+ \sum_{y \in \{0,1\}}
\int_{\mathcal{X}}
  v_{m}^{y}(x)f(x)dx \leq 0 
      \quad\forall m\in[M]
    \right\}.
\end{align}
Therefore, these optimization problems can be expressed as linear programs.
Solving them yields an optimal (randomized) classifier. 
Hence, the proof is complete.
\end{proof}

\subsection{Proof of Theorem \ref{th:Fairbayes_form_MD}}
\begin{proof}
By Lemma \ref{apx:suppt_lem_menon2018}, we have 
\begin{equation}\label{eq:apx_th_bayes_fair_MD1}
R_{cs}(f;c) = (1 - c) \cdot {P}(Y = 1) + {E}_{X}\left[(c - \eta(X)) \cdot f(X)\right],    
\end{equation}
where $\eta(x)=P(Y=1\mid X=x)$.

Recall that $f_{B}^{*}(x)\in\argmin_{f\in\mathcal{F}}\left\{R(f): \mathrm{MD}(f)\leq\delta\right\}$. By Lemma~\ref{lem:Lagrangian_formlation}, there exists some $\{\lambda_{m}\}_{m=1}^{M}$ such that 
\begin{align}\label{eq:apx_th_bayes_fair_MD_probeqv}
\min_{f \in \mathcal{F}} \left\{R_{cs}(f; c) : \mathrm{MD}(f) \leq \delta\right\} = \min_{f \in \mathcal{F}} \left\{R_{cs}(f; c) - \sum_{m=1}^{M} \lambda_m \cdot R^\mathrm{MD}_{m}(f)\right\},    
\end{align}
where
\begin{equation}\label{eq:apx_th_bayes_fair_MD_R}
R^\mathrm{MD}_{m}(f)=\sum_{y \in\{0,1\}} \left\{\Big[  \sum_{m'=1}^{M} a_{m'} b_{m'}^{y} \cdot P\left(\hat{Y}_f=1\mid E_{y,m'}\right)\Big]- {b}_m^y P\left(\hat{Y}_f=1 \mid E_{y,m}\right)+ {c}_m^y \right\}.
\end{equation}

Using Lemma~\ref{apx:suppt_lem_1}, which states that:
$$P(\hat{Y}_{f}=1\mid E_{y,m})={E}_{X}\left[{\gamma^{y}_{m}(X)}\cdot f(X)\right],$$
where $\gamma^{y}_{m}(X)=\frac{P(E_{y,m}\mid X=x)}{P(E_{y,m})}$ and $E_{y,m} = \{Y=y, S=m\}$.
We substitute this result into \eqref{eq:apx_th_bayes_fair_MD_R}, yielding: 
\begin{align}\label{eq:apx_th_bayes_fair_MD2}
R^\mathrm{MD}_{m}(f)
=
  \sum_{y\in\{0,1\}} c_m^y
+
  \sum_{y\in\{0,1\}} {E}_X \left[
    -\left( b_m^y {\gamma_{m}^y(X)} -
    \sum_{m'=1}^M a_{m'}b_{m'}^y {\gamma_{m'}^y(X)}\right)\cdot f(X)
  \right].    
\end{align}
Note that the term $\sum_{y\in\{0,1\}} c_m^y$ is independent of $f$. 
Plugging \eqref{eq:apx_th_bayes_fair_MD1} and \eqref{eq:apx_th_bayes_fair_MD2} into \eqref{eq:apx_th_bayes_fair_MD_probeqv}, the fairness-aware learning problem reduces to: 
\begin{align}\label{eqapx:Lag_problm_MD}
&\min_{f \in \mathcal{F}} \left\{R_{cs}(f; c) - \sum_{m=1}^{M} \lambda_m \cdot R^\mathrm{MD}_{m}(f)\right\}\\\nohao
=&\min_{f \in \mathcal{F}} \left\{ R_{cs}(f; c)- \sum_{m=1}^{M} \lambda_m \cdot 
\sum_{y\in\{0,1\}} {E}_X \left[
    -\left( b_m^y {\gamma_{m}^y(X)} -
    \sum_{m'=1}^M a_{m'}b_{m'}^y {\gamma_{m'}^y(X)}\right)\cdot f(X)
  \right] \right\}\\\nohao
  =&\min_{f \in \mathcal{F}} \left\{ R_{cs}(f; c)-  
\sum_{y\in\{0,1\}} \sum_{m=1}^{M} \lambda_m \cdot {E}_X \left[
    -\left(b_m^y {\gamma_{m}^y(X)} -
    \sum_{m'=1}^M a_{m'}b_{m'}^y {\gamma_{m'}^y(X)}\right)\cdot f(X)
  \right] \right\}\\\nohao
  =&\min_{f \in \mathcal{F}} \left\{ R_{cs}(f; c)-  
  {E}_X\left[
    -\hspace{-5pt}\sum_{y\in\{0,1\}}\hspace{-2pt}\left( \sum_{m=1}^{M} \lambda_m \cdot b_m^y \gamma_{m}^y(X)
    -
    \Lambda_{M} \cdot\hspace{-5pt}\sum_{m'=1}^M a_{m'}b_{m'}^y  {\gamma_{m'}^y(X)}\right)\hspace{-1pt}\cdot f(X)
  \right] \right\}
  \\\nohao
  =&\min_{f \in \mathcal{F}} \left\{ {E}_{X}\left[-(\eta(X)-c) f(X)\right]- \hspace{-1pt} 
  {E}_X \left[
    -\left(\sum_{y\in\{0,1\}} \sum_{m=1}^M
b_m^y\left(\lambda_m - \Lambda_M   a_m
\right) {\gamma_{m}^y(X)}\right)
\hspace{-1pt}\cdot f(X)
  \right] \right\}\\\nohao
  =&\min_{f \in \mathcal{F}} 
  {E}_X \left\{  - \left[\eta(X) - c
    -
      \sum_{m=1}^M\sum_{y\in\{0,1\}} 
        b_m^y\left(\lambda_m - \Lambda_M   a_m
\right)
{\gamma_{m}^y(X)}
  \right]\cdot f(X)
\right\}\\\label{eqapx:Lag_problm_MD_rule}
  =&\min_{f \in \mathcal{F}} 
  {E}_X \left[  - H^{*}_{B}(X)\cdot f(X)
\right],
\end{align}
where $\Lambda_{M}=\sum_{m=1}^{M}\lambda_{m}$ and $$H^{*}_{B}(x)=\eta(X) - c
    - \sum_{m=1}^M\sum_{y\in\{0,1\}} 
        b_m^y\left(\lambda_m - \Lambda_M   a_m
\right) {\gamma_{m}^y(X)}.$$
{Denote $\tilde{f}^{*}_{B}(x)$ the minimizer of problem \eqref{eqapx:Lag_problm_MD}.
According to \eqref{eqapx:Lag_problm_MD_rule}, at optimality, it has a form of $\tilde{f}^{*}_{B}(x) = \mathbbm{1}[H^{*}_{B}(x) > 0]$ when $H^{*}_{B}(x) \neq 0$, and any choice of $\tilde{f}^{*}_{B}(x)$ is admissible when $H^{*}_{B}(x) = 0$. 
In other words, 
\begin{equation}\label{eq:apx_th44_thre}
\tilde{f}^{*}_{B}(x)=\mathbbm{1}\left[H_{B}^{*}(x)>0\right]+\alpha\cdot\mathbbm{1}\left[H_{B}^{*}(x)=0\right].    
\end{equation}
Note that if the uniqueness of the solution holds, then the two optimization problems in \eqref{eq:apx_th_bayes_fair_MD_probeqv} are equivalent, ensuring that ${f}^{*}_{B}(x)=\tilde{f}^{*}_{B}(x)$.
If the solution is not unique, the optimal solution of the Lagrangian form in \eqref{eq:apx_th_bayes_fair_MD_probeqv} contains (at least one) solution(s) from the optimal set of the original constrained problem. 
Thus, there exists ${f}^{*}_{B}(x)$ sharing the same threshold form of $\tilde{f}^{*}_{B}(x)$ as in \eqref{eq:apx_th44_thre}.
This completes the proof.}
\end{proof}

\subsection{Proof of Theorem \ref{th:Fairbayes_form_MR}}

\begin{proof}
By Lemma \ref{apx:suppt_lem_menon2018}, we have 
\begin{equation}\label{eq:apx_th_bayes_fair_MR1}
R_{cs}(f;c) = (1 - c) \cdot {P}(Y = 1) + {E}_{X}\left[(c - \eta(X)) \cdot f(X)\right],  
\end{equation}
where $\eta(x)=P(Y=1\mid X=x)$.

By Lemma \ref{lem:Lagrangian_formlation}, there exists $\{\lambda_{m}\}_{m=1}^{M}$ such that 
\begin{equation}\label{eq:apx_th_bayes_fair_MR_probeqv}
\min_{f \in \mathcal{F}} \left\{R_{cs}(f; c) : \mathrm{MR}(f) \geq \delta\right\} = \min_{f \in \mathcal{F}} \left(R_{cs}(f; c) - \sum_{m=1}^{M} \lambda_m \cdot R^\mathrm{MR}_{m}(f)\right),    
\end{equation}
where
\begin{equation}\label{eq:apx_th_bayes_fair_MR_R}
R^\mathrm{MR}_m(f) =\hspace{-5pt}\sum_{y \in\{0,1\}} \left\{\delta \cdot\Big[ \hspace{-3pt} \sum_{m'=1}^{M} \hspace{-3pt} a_{m'} b_{m'}^{y} \cdot P\left(\hat{Y}_f=1\mid E_{y,m'}\right)\Big] \hspace{-2pt}- {b}_m^y P\left(\hat{Y}_f=1 \mid E_{y,m}\right)+ {c}_m^y \right\}.
\end{equation}

Using Lemma~\ref{apx:suppt_lem_1}, which states that:
$$P(\hat{Y}_{f}=1\mid E_{y,m})={E}_{X}\left[{\gamma^{y}_{m}(X)}\cdot f(X)\right],$$
where $\gamma^{y}_{m}(X)=\frac{P(E_{y,m}\mid X=x)}{P(E_{y,m})}$ and $E_{y,m} = \{Y=y, S=m\}$.
We substitute this result into \eqref{eq:apx_th_bayes_fair_MR_R}, yielding: 
\begin{align}\label{eq:apx_th_bayes_fair_MR2}
R^\mathrm{MR}_m(f) 
=
  \sum_{y\in\{0,1\}} c_m^y
+
  \sum_{y\in\{0,1\}} {E}_X \left[
    -\left( b_m^y {\gamma_{m}^y(X)} -
    \delta\sum_{m'=1}^M a_{m'}b_{m'}^y {\gamma_{m'}^y(X)}\right)\cdot f(X)
  \right]. 
\end{align}

Plugging \eqref{eq:apx_th_bayes_fair_MR1} and \eqref{eq:apx_th_bayes_fair_MR2} into \eqref{eq:apx_th_bayes_fair_MR_probeqv}, the fairness-aware learning problem reduces to: 
\begin{align}\label{eqapx:Lag_problm_MR}
&\min_{f \in \mathcal{F}} \left\{R_{cs}(f; c) - \sum_{m=1}^{M} \lambda_m \cdot R^\text{MR}_{m}(f)\right\}\\\nohao
=&\min_{f \in \mathcal{F}} \left\{ R_{cs}(f; c)- \sum_{m=1}^{M} \lambda_m \cdot \hspace{-5pt}
\sum_{y\in\{0,1\}} {E}_X \left[
    -\left( b_m^y {\gamma_{m}^y(X)} -
    \delta\sum_{m'=1}^M a_{m'}b_{m'}^y {\gamma_{m'}^y(X)}\right)\cdot f(X)
  \right] \right\}\\\nohao
  =&\min_{f \in \mathcal{F}} \left\{ R_{cs}(f; c)-  
  {E}_X \left[
    -\hspace{-5pt}\sum_{y\in\{0,1\}} \hspace{-2pt}\left( \sum_{m=1}^{M} \lambda_m  b_m^y \gamma_{m}^y(X)
    -
    \delta\Lambda_{M} \cdot\hspace{-4pt}\sum_{m'=1}^M a_{m'} 
b_{m'}^y {\gamma_{m'}^y(X)}\right)\hspace{-2pt}\cdot f(X)
  \right] \right\}
  \\\nohao
  =&\min_{f \in \mathcal{F}} \left\{ {E}_{X}\left[-(\eta(X)-c)  f(X)\right]-  \hspace{-1pt}
  {E}_X \hspace{-1pt}\left[
    -\left(\sum_{y\in\{0,1\}} \sum_{m=1}^M
b_m^y\left(
  \lambda_m \hspace{-2pt}   
  -
  \delta\Lambda_M   a_m
\right) 
{\gamma_{m}^y(X)}\hspace{-2pt}\right)\hspace{-2pt}\cdot f(X)
  \right] \right\}\\\nohao
  =&\min_{f \in \mathcal{F}} 
  {E}_X \left\{  - \left[\eta(X) - c
    -
      \sum_{m=1}^M\sum_{y\in\{0,1\}} 
        b_m^y\left(
  \lambda_m   
  -
  \delta\Lambda_M   a_m
\right)
{\gamma_{m}^y(X)}
  \right]\cdot f(X)
\right\}\\\label{eqapx:Lag_problm_MR_rule}
  =&\min_{f \in \mathcal{F}} 
  {E}_X \left[  - H^{*}_{B}(X)\cdot f(X)
\right],
\end{align}
where $\Lambda_{M}=\sum_{m=1}^{M}\lambda_{m}$ and $$H^{*}_{B}(x)=\eta(X) - c
    -
      \sum_{m=1}^M\sum_{y\in\{0,1\}} 
        b_m^y\left(
  \lambda_m   
  -
  \delta\Lambda_M   a_m
\right)
{\gamma_{m}^y(X)}.$$
{Similar to MD case, we denote $\tilde{f}^{*}_{B}(x)$ the minimizer of problem \eqref{eqapx:Lag_problm_MR}.
According to \eqref{eqapx:Lag_problm_MR_rule}, at optimality, it has a form of $\tilde{f}^{*}_{B}(x) = \mathbbm{1}[H^{*}_{B}(x) > 0]$ when $H^{*}_{B}(x) \neq 0$, and any choice of $\tilde{f}^{*}_{B}(x)$ is admissible when $H^{*}_{B}(x) = 0$. 
In other words, 
\begin{equation}\label{eq:apx_MR_th4.5}
\tilde{f}^{*}_{B}(x)=\mathbbm{1}\left[H_{B}^{*}(x)>0\right]+\alpha\cdot\mathbbm{1}\left[H_{B}^{*}(x)=0\right].    
\end{equation}
Note that if the uniqueness of the solution holds, the two optimization problems in \eqref{eq:apx_th_bayes_fair_MR_probeqv} are equivalent.
Then, ${f}^{*}_{B}(x)=\tilde{f}^{*}_{B}(x)$.
If the solution is not unique, the optimal solution of the Lagrangian form in \eqref{eq:apx_th_bayes_fair_MR_probeqv} contains (at least one) solution(s) from the optimal set of the original constrained problem. 
Thus, there exists ${f}^{*}_{B}(x)$ sharing the same threshold form of $\tilde{f}^{*}_{B}(x)$ as in \eqref{eq:apx_MR_th4.5}.}  
This completes the proof.
\end{proof}

\subsection{Proof of Theorem \ref{th:FairCS_BayesOptimal}}

\begin{proof}
By definition,
\begin{align}\nohao
&R_{FCS}^{\bs{\lambda}}(f)
=\sum_{y \in \{0,1\}} \int_{\mathcal{X}} c^{\bs{\lambda}}_{y}(x) P(\hat{Y}_{f} = 1-y, Y = y \mid X = x)   dP_{X}(x)\\\nohao
&=\int_{\mathcal{X}} c^{\bs{\lambda}}_{0}(x) P(\hat{Y}_{f} = 1, Y = 0 \mid X = x)   dP_{X}(x)+\int_{\mathcal{X}} c^{\bs{\lambda}}_{1}(x) P(\hat{Y}_{f} = 0, Y = 1 \mid X = x)   dP_{X}(x)\\\nohao
&=\int_{\mathcal{X}} c^{\bs{\lambda}}_{0}(x) P(\hat{Y}_{f} = 1, Y = 0 \mid X = x)   dP_{X}(x)\\\nohao
&\qquad+\int_{\mathcal{X}} \left(1-c^{\bs{\lambda}}_{0}(x)\right) P(\hat{Y}_{f} = 0, Y = 1 \mid X = x)   dP_{X}(x)\\\nohao
&=\int_{\mathcal{X}} c^{\bs{\lambda}}_{0}(x) 
f(x)[1-\eta(x)]   dP_{X}(x)+\int_{\mathcal{X}} \left(1-c^{\bs{\lambda}}_{0}(x)\right)\eta(x)[1-f(x)]   dP_{X}(x)\\
&=\int_{\mathcal{X}}
-f(x) \left[
\eta(x)-c^{\bs{\lambda}}_{0}(x)
\right]
 dP_X(x)+\int_{\mathcal{X}}
\left[
  \left(1 - c^{\bs{\lambda}}_{0}(x)\right) \eta(x)
\right]
 dP_X(x).
\end{align}    
Note that the second term does not depend on $f$ and that the first term is minimized by taking $f(x)=1$ if $\eta(x)-c^{\bs{\lambda}}_{0}(x)>0$ and $f(x)=0$ if $\eta(x)-c^{\bs{\lambda}}_{0}(x)<0$.
Therefore, we can conclude that, for all \(x\),
\begin{equation}\label{eq:apx_fBIn}
f_{B}^\text{In}(x)=\mathbbm{1}\left[\eta(x)-c^{\bs{\lambda}}_{0}(x)>0\right].    
\end{equation}
Here, the cost \(c^{\bs{\lambda}}_{0}(x)\) is given as follows according to its definition:
\begin{itemize}[leftmargin=2em]
    \item For the MD measure:
    \[
    c^{\bs{\lambda}}_{0}(x) = c +  \sum_{m=1}^M \sum_{y \in \{0,1\}}
    b_m^y\left(\lambda_m -\Lambda_M a_m\right)\gamma_m^y(x).
    \]
    \item For the MR measure:
    \[
    c^{\bs{\lambda}}_{0}(x) = c + \sum_{m=1}^M \sum_{y \in \{0,1\}}
    b_m^y\left(\lambda_m -\delta\Lambda_M a_m\right)\gamma_m^y(x).
    \]
\end{itemize}

Thus, \eqref{eq:apx_fBIn} recovers the Bayes-optimal fair classifiers as shown in Theorems~\ref{th:Fairbayes_form_MD} and~\ref{th:Fairbayes_form_MR}, which completes the proof.
\end{proof}

\subsection{Fair Cost-Sensitive Classifier with \textit{S}}\label{apxsec:C.8}
When $S$ is available for prediction, the construction of the Bayes-optimal fair cost-sensitive classifier simplifies, as shown in the following corollary.

\begin{corollary}[Fair Cost-Sensitive Classifier with $ S $]\label{corl:FairCS_BayesOptimal_S}
Let $ c^{\bs{\lambda}}_{y}(x, s) = (1 - 2y)\left[c + Q^{\bs{\lambda}}(x, s)\right] + y $, where $ y \in \{0, 1\} $ and:
{\begin{align}\nohao
    Q_\mathrm{MD}^{\bs{\lambda}}(x, s) &=\sum_{y\in\{0,1\}}b_{s}^{y}\left(\lambda_{s}-\Lambda_{M}a_{s}\right){\gamma_{s}^{y}(x,s)},
    \\\nohao
    Q_\mathrm{MR}^{\bs{\lambda}}(x, s) &= \sum_{y\in\{0,1\}}b_{s}^{y}\left(\lambda_{s}-\delta\cdot\Lambda_{M}a_{s}\right){\gamma_{s}^{y}(x,s)}.
\end{align}
}Here, $ \bs{\lambda} $, $ \lambda_{s} $, $ \Lambda_{M} $, $ \gamma_{s}^{y}(x,s) $, and the values of $ a_{s} $ and $ b_{s}^{y} $ are as defined in Corollary \ref{corl:BayesClass_withS_MD}.

Define the fair cost-sensitive risk of a classifier $ f $ as:
{\begin{align}\label{eq:fair_CS_risk_S}
   R_{FCS}^{\bs{\lambda}}(f) = \sum_{y \in \{0, 1\}} \sum_{s = 1}^{M} \left\{\int_{\mathcal{X}} c^{\bs{\lambda}}_{y}(x, s) \cdot P(\hat{Y}_{f} = 1-y, Y = y \mid X = x, S = s) \, dP_{X, S}(x, s)\right\}.
\end{align}}Then, $ f_{B}^\text{In}(x, s) = \argmin_{f \in \mathcal{F}} R_{FCS}^{\bs{\lambda}}(f) $ is the Bayes-optimal fair classifier.
\end{corollary}

To illustrate Corollary \ref{corl:FairCS_BayesOptimal_S}, we take MD as an example.
This corollary implies that when $S$ is available for prediction:
(1) For negative examples in group $ s $, the weight is $c_{0}^{\bs{\lambda}}(x, s) = c + \sum_{y\in\{0,1\}}b_{s}^{y}\left(\lambda_{s}-\Lambda_{M}a_{s}\right){\gamma_{s}^{y}(x,s)}$ 
and (2) for positive examples in group $ s $, the weight is $c_{1}^{\bs{\lambda}}(x, s) = 1 - c_{0}^{\bs{\lambda}}(x, s).$
The fair cost-sensitive classifier is then trained by minimizing \eqref{eq:fair_CS_risk_S} on $\{x_i,s_i,y_i\}_{i=1}^{N}$ instead of $\{x_i,y_i\}_{i=1}^{N}$ as specified in line 4 of Algorithm~\ref{algo:in_processing}.

{ 
}

\section{Experiments}\label{apxsex:Experiments}

\subsection{Datasets}\label{apxsex:Experiments_Datasets}
We conduct experiments on two real-world classification benchmark datasets:
(1) \texttt{Adult} \citep{adult_2}:  A UCI dataset where the task is to predict whether the income of an individual is over $\$50$k per year; 
(2) \texttt{COMPAS} \citep{angwin2016machinebias}: A dataset where the task is to predict the recidivism of criminals. 
Both datasets are used for binary classification tasks.
They also contain demographic features such as \textit{gender} (binary) and \textit{race} (binary). 
We use these features to construct scenarios with a single sensitive feature as well as multiple sensitive features.

For the \texttt{Adult} dataset, we first consider \textit{gender} as the single (binary) sensitive feature to evaluate performance in a conventional fairness setting.  
We then extend the analysis to the case of  multiple sensitive features by incorporating both \textit{race}  and \textit{gender} simultaneously.  
For \textit{race} , we select the two largest racial subgroups and treat it as a binary variable.  
This binary \textit{race}  variable is then combined with \textit{gender} to define intersectional subgroups, resulting in four non-overlapping subgroups for final experimental evaluations.

Similarly, for the \texttt{COMPAS} dataset, we consider \textit{race}  as a binary sensitive feature, selecting ``African-American'' and ``Caucasian'' groups.  
We then extend the analysis to multiple sensitive features by incorporating both \textit{race}  and \textit{gender} simultaneously.  
Detailed dataset statistics are provided in Table \ref{tab:datasets}.  
Each dataset is split into training and testing sets in a 0.5:0.5 ratio.
\begin{table}[htp]

\begin{center}

\begin{tabular}{lllc}
\toprule
Dataset & $(N, d)$ & Sensitive Features & $M$ \\
\midrule
\texttt{Adult} (single) & $(48842, 108)$ &  \textit{gender}  & 2 \\
\texttt{COMPAS} (single) & $(5278, 7)$    &    \textit{race} & 2 \\                                     
\texttt{Adult} (multiple) & $(15507, 94)$ &  \textit{gender} \&   \textit{race}   & 4 \\
\texttt{COMPAS} (multiple) & $(5278, 7)$    &   \textit{gender} \& \textit{race} & 4 \\ 
\bottomrule
\end{tabular}

\end{center}
\caption{Datasets Details ($N$: dataset size, $d$: dimension of $X$, $M$: the number of subgroups).}
\label{tab:datasets}
\end{table}

\subsection{Algorithms Setup \& Hyperparameters}
\label{appendix:Setup}
\setcounter{footnote}{0}

We further split the training data into two equal parts, using one half for model training and the remaining half for fine-tuning the parameters. 
{ We first train predictors on the training data to estimate the probabilities $ \Tilde{P}(Y  \mid X = x) $  and $ \Tilde{P}(S, Y \mid X = x)$, which are then incorporated into the proposed algorithms. 
For the \texttt{COMPAS} dataset with a binary sensitive feature, we use LR as the predictor for these probability estimations, while for other settings and datasets, we use XGBoost.}

As both our method and the baseline \textit{LinearPost} \citep{xian2024unifiedpostprocessing} require training a base predictor to model the joint distribution of the target and sensitive features, we use the same base predictor for both methods to ensure a fair comparison. 
{ We include both calibrated and uncalibrated versions of \textit{LinearPost}\footnote{\url{https://github.com/uiuctml/fair-classification}} for comparison.} 
For the \textit{Reduction} \citep{agarwal2018reductions} algorithm, we use the implementation from \textit{fairlearn} library \citep{Weerts2023}, with its exponentiated gradient variants.
We also include \textit{MBS}\footnote{\url{https://github.com/chenw20/BiasScore}} \citep{chen2024posthoc} as a fair baseline.
{ In addition, we include a basic baseline classifier without fairness constraints for comparison.  }

Furthermore, following \citet{menon2018} and \citet{chen2024posthoc}, the values of the hyperparameter $ \bs{\lambda} \in  [-1,1]^{M}$ are selected via grid search; a step size of $0.01$ is used.
It is noteworthy that Algorithm \ref{algo:post_processing} does not involve re-training a classifier, making the search for $\bs{\lambda}$ substantially faster compared to Algorithm \ref{algo:in_processing}.
For various values of $\bs{\lambda}$, we report both accuracy and fairness levels, as they represent the key performance metrics of interest. 
All four fairness notions in Table~\ref{tab:genernal_fair_measure} are considered to show the generality of our framework.  
Approximate fairness is implemented using both MD and MR measures, and the achieved fairness level (i.e., values of $\mathrm{MD}(f)$ and $\mathrm{MR}(f)$) are reported.

{All experiments were conducted on a Mac Studio equipped with an M2 Max chip and 64GB of memory, implemented in Python 3.8. 
Our code will be publicly available upon acceptance.
}


\subsection{Extension Results}\label{apx:expresults}

\subsubsection{Results on DP and EO notions}

{Figure \ref{apxfig:combined_data_fairness} shows the fairness-accuracy trade-offs for MD measure under DP and EO across two datasets. 
Note that all fair baselines used in our experiments support both of these two notions.
For DP and EO, compared to the fair baselines, our algorithms achieve the more favorable fairness-accuracy trade-off in most cases, especially in the attribute-blind setting. 
Notably, for all fair methods, their performance gap between attribute-blind and attribute-aware settings is larger on \texttt{COMPAS}. 
This is likely due to its smaller dataset size, which may hinder accurate estimation of $P(Y|X)$ and $P(S, Y|X)$. 
Additionally, our in-processing method often outperforms our post-processing one in balancing fairness and accuracy, with a more significant advantage on the \texttt{COMPAS} dataset.}
   
\begin{figure*}[!htbp]
 \centering
 \includegraphics[width=\textwidth]{figs/legend_R3.pdf} 
\begin{center} 
   \begin{subfigure}{0.49\textwidth}
        \centering
        \begin{subfigure}{0.49\textwidth} 
            \centering
            \includegraphics[width=\linewidth]{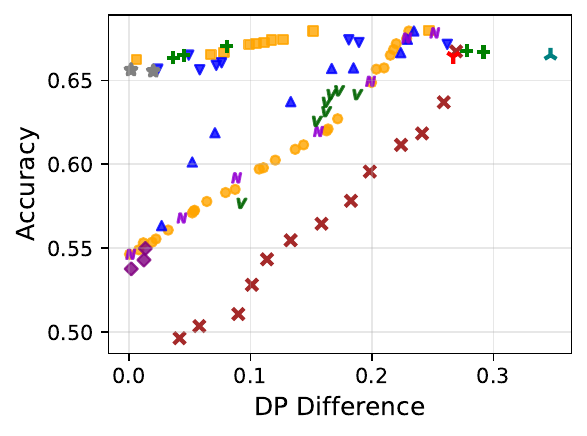}
        \end{subfigure}
        \begin{subfigure}{0.49\textwidth} 
            \centering
            \includegraphics[width=\linewidth]{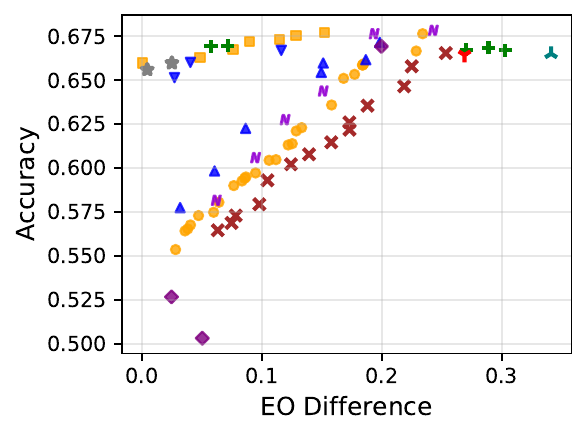}
        \end{subfigure}
        \begin{subfigure}{0.49\textwidth} 
            \centering
            \includegraphics[width=\linewidth]{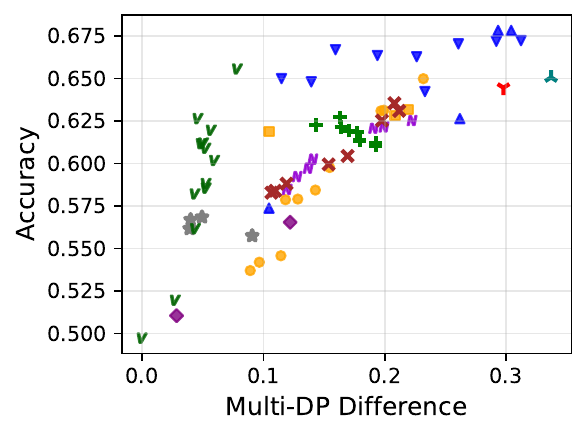}
        \end{subfigure}
        \begin{subfigure}{0.49\textwidth} 
            \centering
            \includegraphics[width=\linewidth]{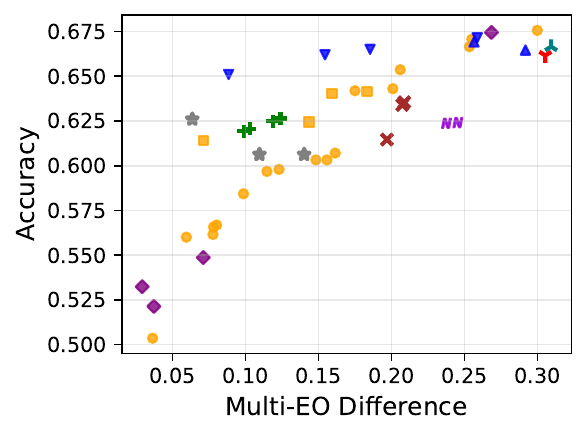}
        \end{subfigure}
        \caption{\texttt{COMPAS}}
        \label{apxfig:compas}
    \end{subfigure}
    \hfill
    \begin{subfigure}{0.49\textwidth}
        \centering
        \begin{subfigure}{0.49\textwidth} 
            \centering
            \includegraphics[width=\linewidth]{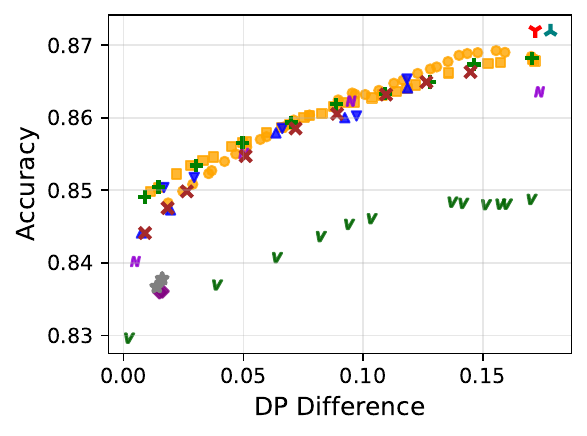}
        \end{subfigure}
        \begin{subfigure}{0.49\textwidth} 
            \centering
            \includegraphics[width=\linewidth]{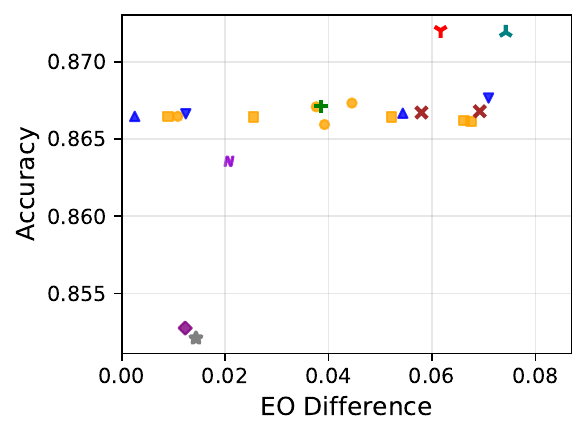}
        \end{subfigure}
        \begin{subfigure}{0.49\textwidth} 
            \centering
            \includegraphics[width=\linewidth]{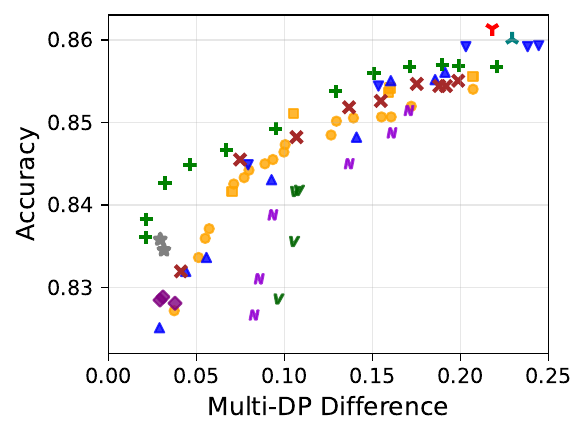}
        \end{subfigure}
        \begin{subfigure}{0.49\textwidth} 
            \centering
            \includegraphics[width=\linewidth]{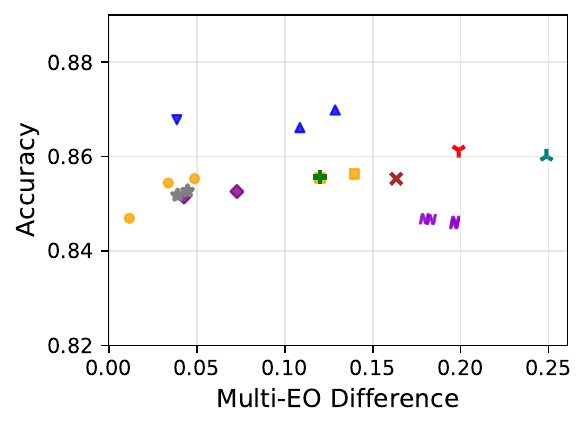}
        \end{subfigure}
        \caption{\texttt{Adult}}
        \label{apxfig:adult}
    \end{subfigure}
            
    \caption{Trade-offs between accuracy and fairness using MD. 
    The prefix `Multi-' represents the case of multiple sensitive features, and `(aware)' in classifier  names indicates the attribute-aware setting. (uncal): uncalibrated; (cal): calibrated.}
    \label{apxfig:combined_data_fairness}
    \end{center}
\end{figure*}

\subsubsection{Results on PE and AP notions}

Figure \ref{fig:two_dataset_PE} presents the results for MD measure under PE and AP across two datasets.
Since no fair baselines explicitly support these fairness notions, we compare our methods to  unconstrained basic classifiers.
The results show that our methods effectively reduce the False Positive Rate gap (for PE) and accuracy gap (for AP) among groups, respectively.
For instance, under PE on the \texttt{Adult} dataset, our methods lower the unfairness level from 0.058 (accuracy 0.872, unconstrained classifiers) to below 0.01 while maintaining an accuracy of approximately 0.861.
Similar results are also observed for AP.
Additionally, Figure \ref{fig:two_dataset_PE} illustrates that adjusting $\bs{\lambda}$ allows for a flexible trade-off between fairness and overall accuracy.

\begin{figure}[!ht]
\begin{center} 
    \begin{subfigure}{\textwidth} 
        \centering
        \includegraphics[width=0.6\textwidth]{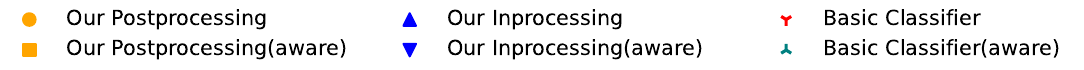}
        
        \begin{subfigure}{0.23\textwidth}
            \centering
            \includegraphics[width=\linewidth]{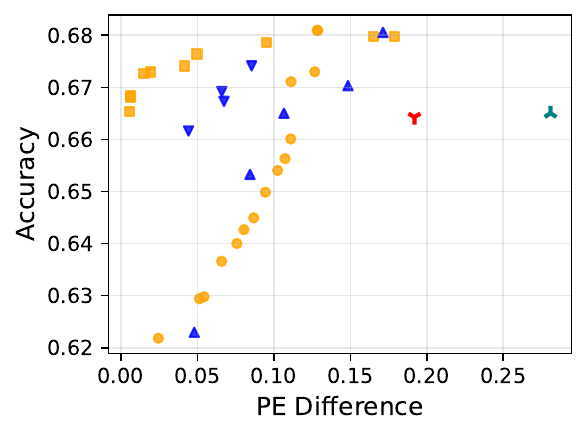}
            \caption{\texttt{COMPAS} / PE}
        \end{subfigure}
        \begin{subfigure}{0.23\textwidth}
            \centering
            \includegraphics[width=\linewidth]{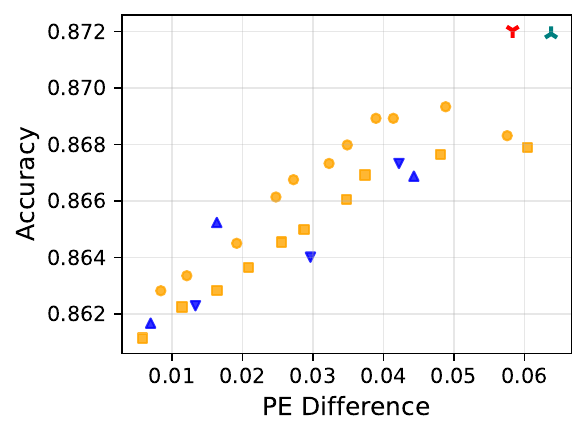}
            \caption{\texttt{Adult} / PE}
        \end{subfigure}
        \begin{subfigure}{0.23\textwidth}
            \centering
            \includegraphics[width=\linewidth]{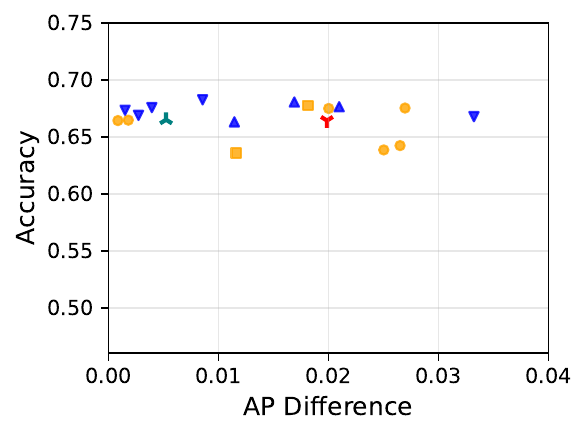}
            \caption{\texttt{COMPAS} / AP}
        \end{subfigure}
        \begin{subfigure}{0.23\textwidth}
            \centering
            \includegraphics[width=\linewidth]{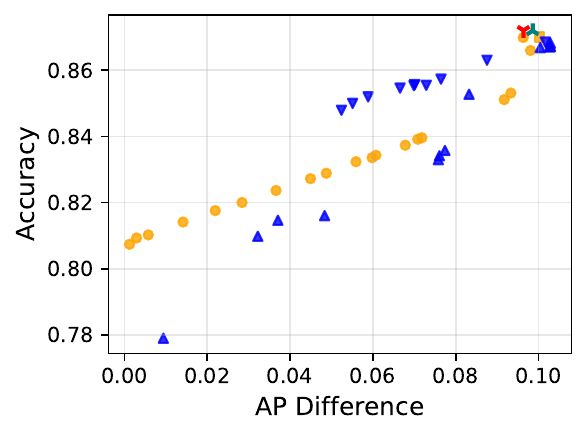}
            \caption{\texttt{Adult} / AP}
        \end{subfigure}
    \end{subfigure}
    \caption{Trade-offs between accuracy and fairness under PE and AP on two datasets. 
    The `(aware)' in the name of classifiers indicates the attribute-aware setting.}
    \label{fig:two_dataset_PE}
    \end{center}
    \vskip -0.1in
\end{figure}

\subsubsection{Results on Mean Ratio Measure}

Figures \ref{fig:compas_fairness} and \ref{fig:adult_MR} present the results for MR measure on \texttt{COMPAS} and \texttt{Adult} datasets.
Again, we compare our methods against unconstrained classifiers, since no fair baselines support the MR measure. 
The results show that our methods effectively increase the ratio of  relevant quantities among groups (thus, reduce the unfairness level).
Again, our in-processing method often achieve favorable performance compared to our post-processing method on \texttt{COMPAS}.

\begin{figure}[!htbp]
\begin{center} 
    \begin{subfigure}{\textwidth} 
        \centering
        \includegraphics[width=0.6\textwidth]{figs/legend_wide_nobase.pdf}
        
        \begin{subfigure}{0.23\textwidth}
            \centering
            \includegraphics[width=\linewidth]{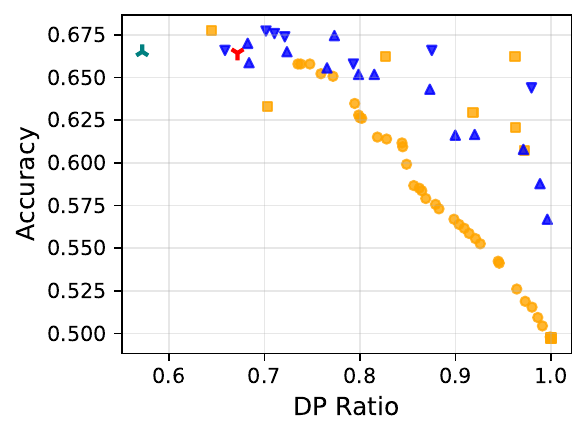}
        \end{subfigure}
        \begin{subfigure}{0.23\textwidth}
            \centering
            \includegraphics[width=\linewidth]{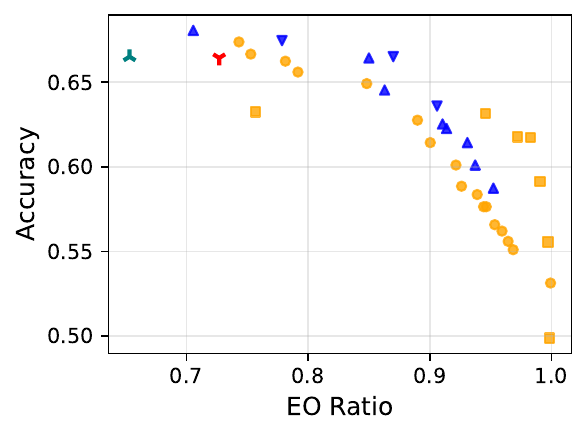}
        \end{subfigure}
        \begin{subfigure}{0.23\textwidth}
            \centering
            \includegraphics[width=\linewidth]{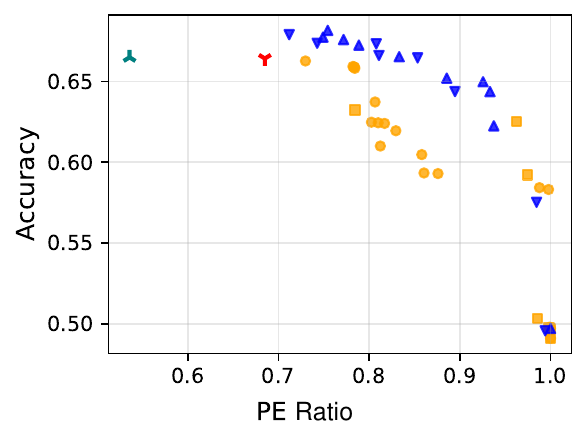}
        \end{subfigure}
    
    \end{subfigure}
    \caption{Trade-offs between accuracy and fairness on \texttt{COMPAS}, using MR measure.
    The `(aware)' in the name of classifiers indicates the attribute-aware setting.}
    \label{fig:compas_fairness}
    \end{center}
    \vskip -0.1in
\end{figure}

\begin{figure}[!htbp]
 
\begin{center} 
    \begin{subfigure}{\textwidth} 
        \centering
        \includegraphics[width=0.6\textwidth]{figs/legend_wide_nobase.pdf}
        
        \begin{subfigure}{0.23\textwidth}
            \centering
            \includegraphics[width=\linewidth]{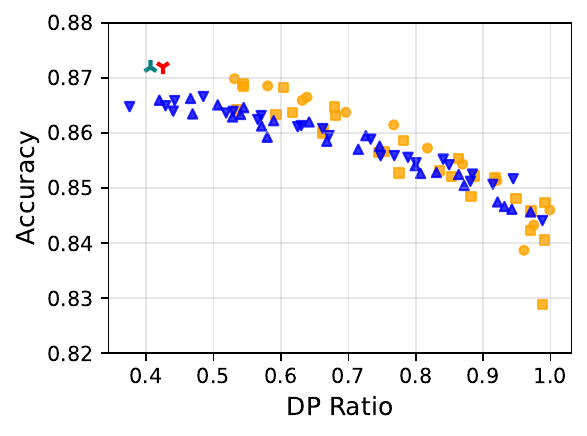}
        \end{subfigure}
        \begin{subfigure}{0.23\textwidth}
            \centering
            \includegraphics[width=\linewidth]{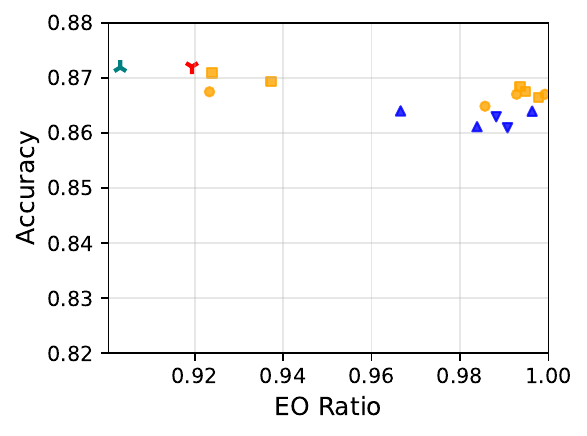}
        \end{subfigure}
        \begin{subfigure}{0.23\textwidth}
            \centering
            \includegraphics[width=\linewidth]{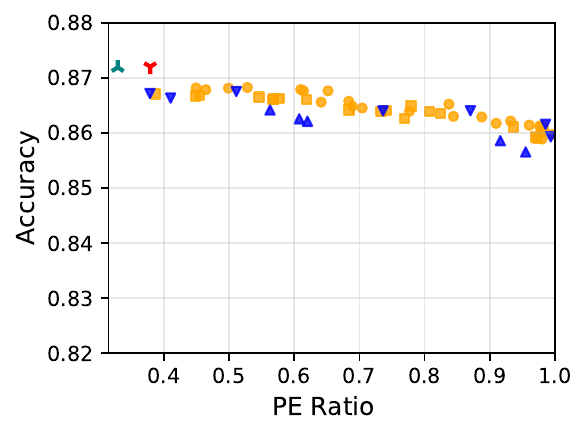}
        \end{subfigure}
    
    \end{subfigure}
    \caption{Trade-offs between accuracy and fairness on \texttt{Adult}, using MR measure.
    The `(aware)' in the name of classifiers indicates the attribute-aware setting.}
    \label{fig:adult_MR}
    \end{center}
    \vskip -0.1in
\end{figure}


{ 

\subsection{Final $\lambda$ Values}

Tables \ref{apxtab:lambdaCOM} and \ref{apxtab:lambdaADT} provide the example selected $\lambda$ values in our experiments.

\begin{table}[!htbp]
\begin{center}
\begin{tabular}{lcccccc}
\toprule
Method                        & MD & Accuracy & $\lambda_1$ & $\lambda_2$ & $\lambda_3$ & $\lambda_4$ \\
\midrule
Our Inprocess & 0.1045              & 0.5737   & -0.25     & -0.9167   & -0.3333   & -0.8333   \\
Our Inprocess & 0.2617              & 0.6264   & 0.0       & 0.9167    & 0.6667    & -0.5      \\
Our Inprocess & 0.2932              & 0.6783   & 0.6667    & -0.8333   & 0.8333    & 0.25 \\    
Our Inprocess (aware) & 0.1147              & 0.6499   & -0.0625   & -0.1875   & -0.0625   & -0.1875   \\
Our Inprocess (aware) & 0.1938              & 0.6635   & -0.125    & -0.5625   & -0.125    & -0.375    \\
Our Inprocess (aware) & 0.3121              & 0.6722   & -0.125    & -0.625    & -0.125    & -0.375   \\
\hline
Our Postprocessing        & 0.0890  & 0.5369   & 0.2         & 0.4         & 0.4         & -0.6        \\
Our Postprocessing        & 0.1282  & 0.5790   & 0.0         & 0.4         & 0.4         & -0.2        \\
Our Postprocessing        & 0.2316  & 0.6499   & 0.2         & 1.0         & 0.4         & 0.8         \\
Our Postprocessing (aware) & 0.1046  & 0.6188   & 0.0         & 0.1         & 0.0         & 0.0         \\
Our Postprocessing (aware) & 0.2084  & 0.6283   & -0.1        & -0.5        & -0.1        & -0.3        \\
Our Postprocessing (aware) & 0.2192  & 0.6317   & 0.0         & 0.0         & 0.0         & 0.0         \\
\bottomrule
\end{tabular}
\end{center}
\caption{Selected  $\lambda$ values on \texttt{COMPAS} for Multi-DP with MD in one random run.}\label{apxtab:lambdaCOM} 
\end{table}

\begin{table}[!htbp]
\begin{center}
\begin{tabular}{lcccccc}
\toprule
Method                 & MD      & Accuracy & $\lambda_1$ & $\lambda_2$ & $\lambda_3$ & $\lambda_4$ \\
\midrule
Our Inprocess          & 0.0290  & 0.8251   & 0.0         & 0.3         & 0.3         & 0.8         \\
Our Inprocess          & 0.0556  & 0.8336   & 0.0         & 0.4         & 0.1         & 0.4         \\
Our Inprocess          & 0.1604  & 0.8550   & 0.3         & 0.0         & -0.2        & -0.4        \\
Our Inprocess (aware)   & 0.0794  & 0.8449   & 0.0         & 0.0         & 0.0         & 0.2         \\
Our Inprocess (aware)   & 0.1534  & 0.8544   & 0.0         & 0.0         & 0.1         & 0.1         \\
Our Inprocess (aware)   & 0.2382  & 0.8592   & 0.1         & 0.0         & 0.1         & 0.0         \\
\hline
Our Postprocessing        & 0.0712  & 0.8425   & -0.1        & 0.2         & -0.4        & -0.8        \\
Our Postprocessing        & 0.1552  & 0.8507   & 0.2         & -0.1        & -0.4        & -0.8        \\
Our Postprocessing        & 0.2072  & 0.8540   & 0.3         & 0.0         & -0.2        & -0.4        \\
Our Postprocessing (aware) & 0.0703  & 0.8416   & 0.0         & 0.0         & -0.1        & 0.1         \\
Our Postprocessing (aware) & 0.1592  & 0.8536   & 0.0         & 0.0         & 0.1         & 0.1         \\
Our Postprocessing (aware) & 0.2071  & 0.8556   & 0.0         & 0.0         & 0.0         & -0.2        \\
\bottomrule
\end{tabular}
\end{center}
\caption{Selected  $\lambda$ values on \texttt{Adult} for Multi-DP with MD in one random run.}\label{apxtab:lambdaADT} 
\end{table}

\subsection{Confidence Interval of Results}

We use \texttt{COMPAS} as a representative example and report numerical results (means and confidence intervals, CI) in Table \ref{apxtab:CI}.

\begin{table}[H]
\begin{center}
\begin{tabular}{lcccc}
\toprule
Method                    & Accuracy & Acc-CI                     & MD       & MD-CI                     \\
\midrule
Our Postprocessing         & 0.5936   & (0.5822, 0.6050)           & 0.1096   & (0.1069, 0.1123)          \\
Our Postprocessing         & 0.6233   & (0.6132, 0.6334)           & 0.1596   & (0.1575, 0.1617)          \\
Our Postprocessing         & 0.6461   & (0.6328, 0.6594)           & 0.2108   & (0.2098, 0.2118)          \\
Our Postprocessing         & 0.6650   & (0.6517, 0.6782)           & 0.2571   & (0.2521, 0.2622)          \\
Our Postprocessing         & 0.6735   & (0.6619, 0.6851)           & 0.2927   & (0.2882, 0.2972)          \\
\hline
Our Postprocessing (aware) & 0.6413   & (0.6256, 0.6569)           & 0.0916   & (0.0722, 0.1110)          \\
Our Postprocessing (aware) & 0.6419   & (0.6291, 0.6547)           & 0.1403   & (0.1279, 0.1527)          \\
Our Postprocessing (aware) & 0.6408   & (0.6318, 0.6498)           & 0.2038   & (0.1968, 0.2107)          \\
Our Postprocessing (aware) & 0.6438   & (0.6258, 0.6618)           & 0.2387   & (0.2289, 0.2484)          \\
Our Postprocessing (aware) & 0.6569   & (0.6351, 0.6787)           & 0.2877   & (0.2811, 0.2943)          \\
\bottomrule
\end{tabular}
\end{center}
\caption{$95\%$ confidence intervals ($t$ distribution, 10 runs) on \texttt{COMPAS} for Multi-DP with MD.}\label{apxtab:CI}
\end{table}

\section{Additional Experiments}\label{apxsex:Additional_Experiments}

\subsection{Larger-Scale Datasets}

We also test our algorithms on three larger-scale data sets from \citet{ding2021retiring}.
The results, shown in Figure~\ref{fig:combined_acs_md}, exhibit trends similar to those observed on the previous datasets.
\begin{figure}[htbp]
 \vskip 0.1in 
 \centering \includegraphics[width=\textwidth]{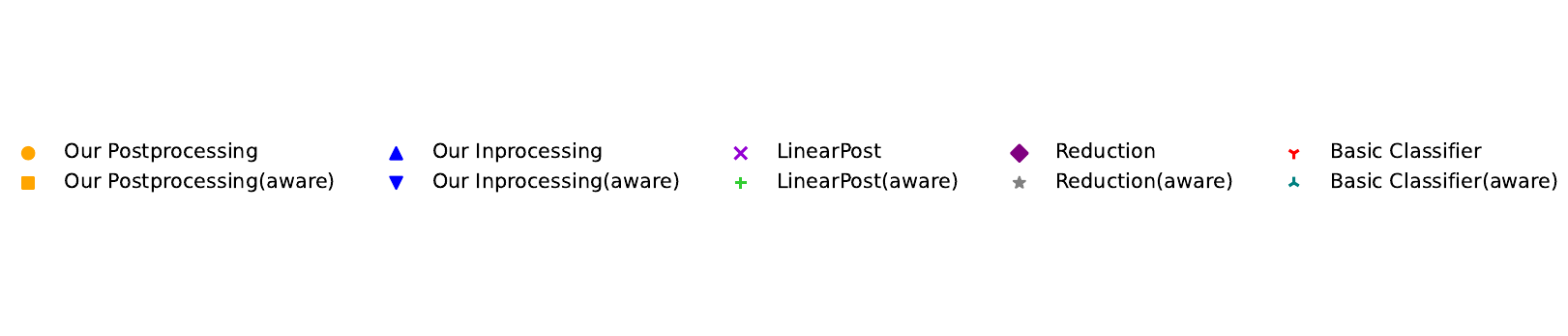}
 \vspace{-4.5em}
\begin{center} 
    \begin{subfigure}{0.32\textwidth}
        \centering
        \begin{subfigure}{\linewidth}
            \centering
            \includegraphics[width=\linewidth]{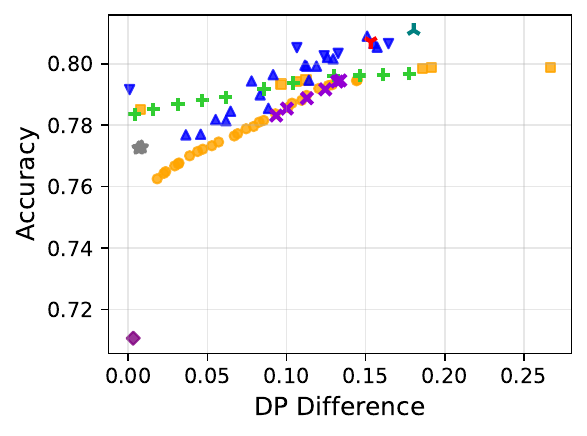}
        \end{subfigure}
        \vskip 1pt
        \begin{subfigure}{\linewidth}
            \centering
            \includegraphics[width=\linewidth]{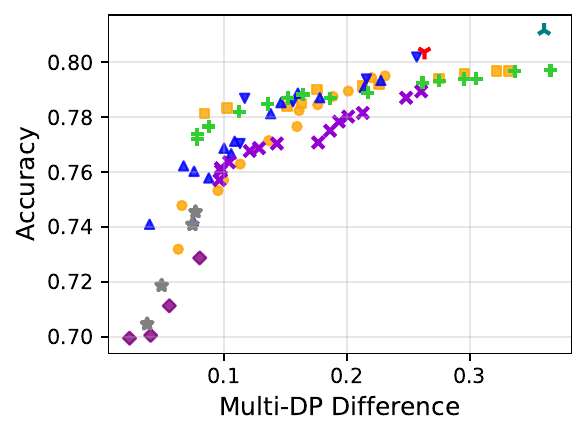}
        \end{subfigure}
        \caption{\texttt{ACS Income}}
        \label{fig:acs_income}
    \end{subfigure}
    \hfill
    \begin{subfigure}{0.32\textwidth}
        \centering
        \begin{subfigure}{\linewidth}
            \centering
            \includegraphics[width=\linewidth]{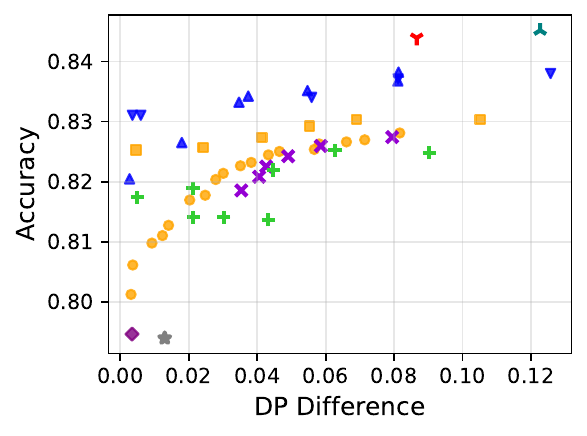}
        \end{subfigure}
        \vskip 1pt
        \begin{subfigure}{\linewidth}
            \centering
            \includegraphics[width=\linewidth]{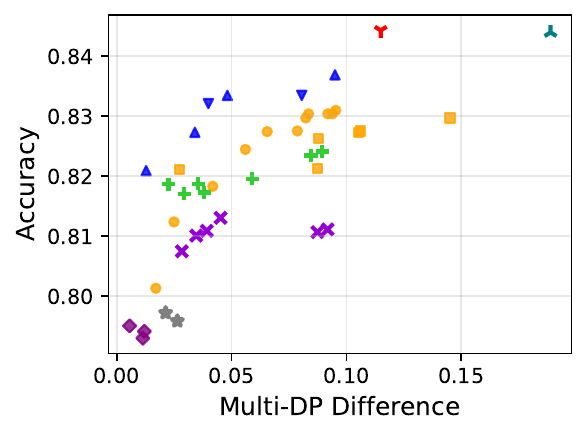}
        \end{subfigure}
        \caption{\texttt{ACS Public Coverage}}
        \label{fig:acs_pubcov}
    \end{subfigure}
    \hfill
    \begin{subfigure}{0.32\textwidth}
        \centering
        \begin{subfigure}{\linewidth}
            \centering
            \includegraphics[width=\linewidth]{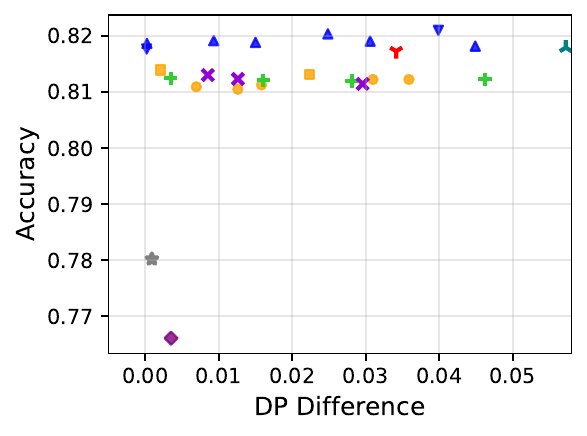}
        \end{subfigure}
        \vskip 1pt
        \begin{subfigure}{\linewidth}
            \centering
            \includegraphics[width=\linewidth]{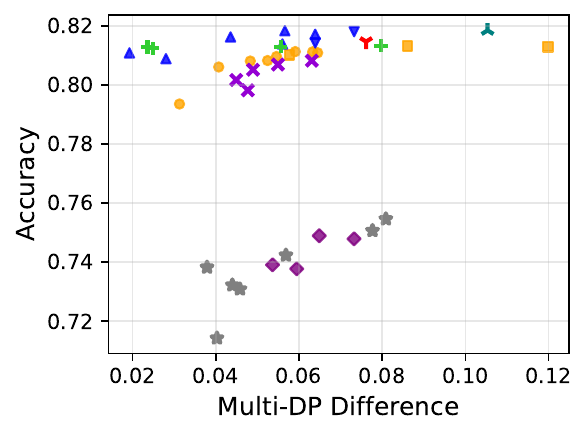}
        \end{subfigure}
        \caption{\texttt{ACS Employment}}
        \label{fig:acs_employment}
    \end{subfigure}
    
    \caption{Accuracy–fairness (DP) trade-offs on additional datasets using MD. 
Top: single sensitive feature (\textit{Race}). 
Bottom (prefix ‘Multi-’): multiple sensitive features (\textit{Race} \& \textit{Sex} combined).  
‘ (aware)’ indicates the attribute-aware setting.}
    \label{fig:combined_acs_md}
    \end{center}
 \vskip -0.1in
\end{figure}

\subsection{Ablations Analysis}

In this ablation analysis, we manually perturb $\hat{P}(Y \mid X,S)$ with noise drawn i.i.d. from a uniform distribution $\mathrm{Unif}(-\epsilon, 2\epsilon)$, where $\epsilon$ controls the corruption intensity.
We use our DP postprocessing method in the attribute-aware setting as a representative example. 
In Figure~\ref{apxfig:ablations}, we plot the accuracy and the DP MD level of our method on the corrupted estimator for the \texttt{COMPAS} dataset. Across all noise intensities, the MD level remains below or fluctuates around the target limit $\delta$, while inuring a minor drop in accuracy.
This demonstrates the robustness of our methods even when the estimated probability do not match the ground-truth well.

\begin{figure}[htbp] 
\begin{center} 
    \begin{subfigure}{\textwidth} 
        \centering
        
        \begin{subfigure}{0.45\textwidth}
            \centering
            \includegraphics[width=\linewidth]{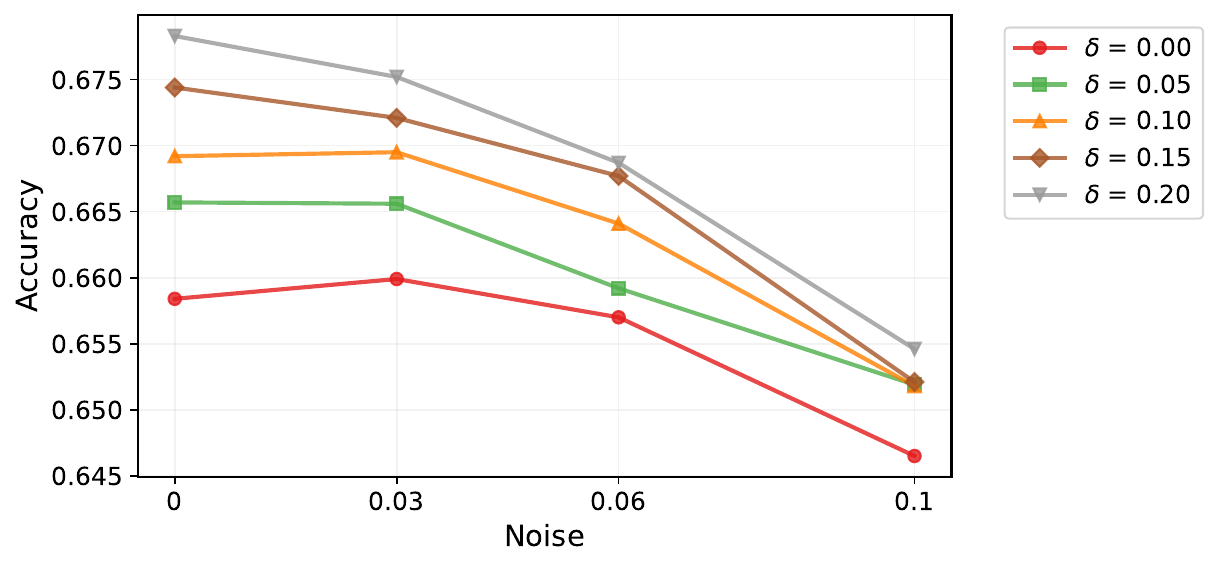}
            \caption{\texttt{COMPAS} / DP}
        \end{subfigure}
        \begin{subfigure}{0.45\textwidth}
            \centering
            \includegraphics[width=\linewidth]{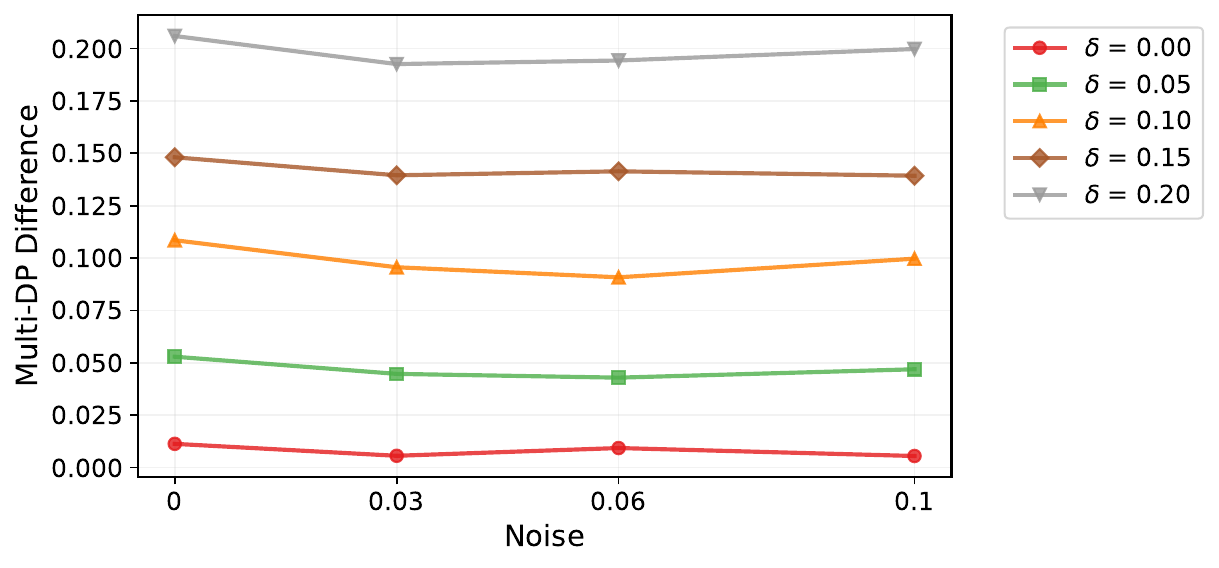}
            \caption{\texttt{COMPAS} / DP}
        \end{subfigure}
        \begin{subfigure}{0.45\textwidth}
            \centering
            \includegraphics[width=\linewidth]{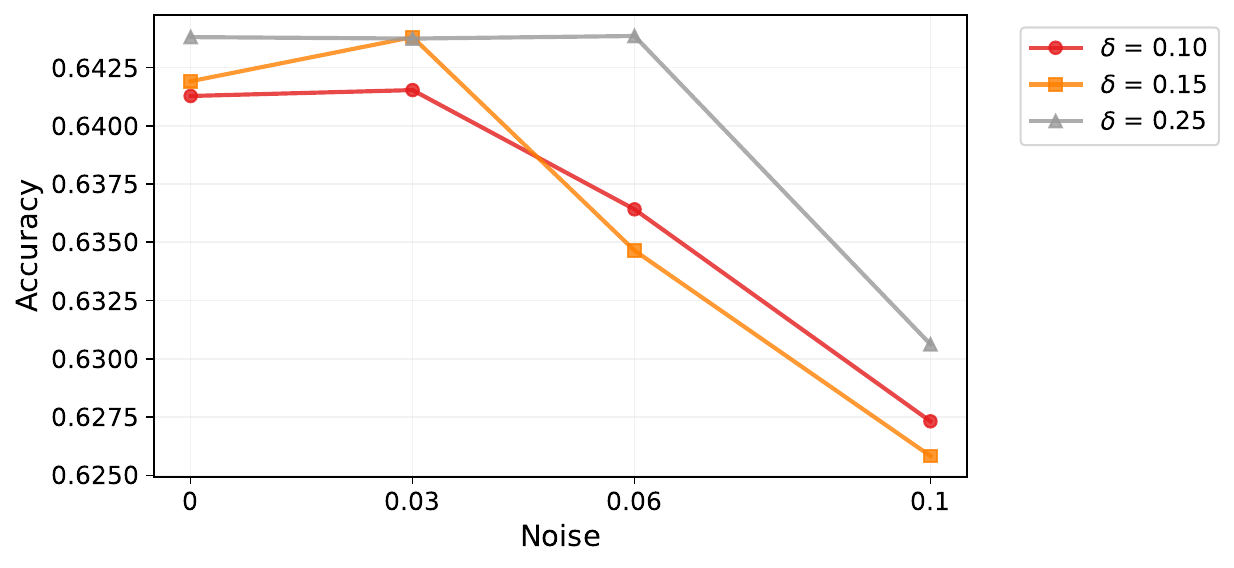}
            \caption{\texttt{COMPAS} / Multi-DP}
        \end{subfigure}
        \begin{subfigure}{0.45\textwidth}
            \centering
            \includegraphics[width=\linewidth]{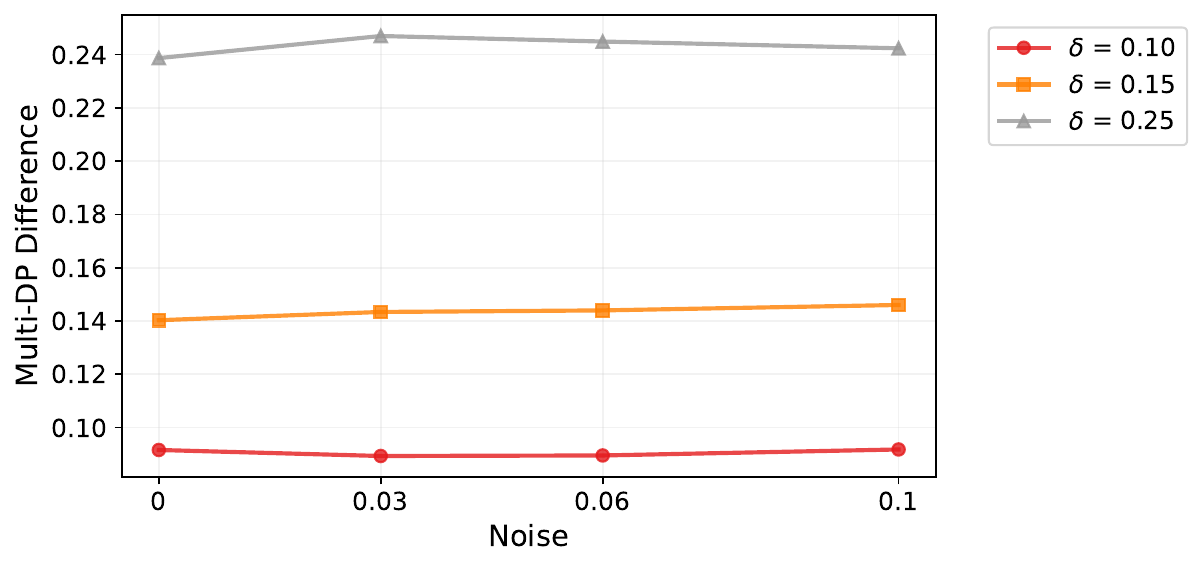}
            \caption{\texttt{COMPAS} / Multi-DP}
        \end{subfigure}
    \end{subfigure}
    \caption{Performance of our  method with corrupted
    estimator on \texttt{COMPAS}.
    }
    \label{apxfig:ablations}
    \end{center}
    \vskip -0.1in
\end{figure}

\subsection{Calibration}\label{apxsec: Calibration}

For group membership probability estimation, the predictions $\tilde{P}(E_{y,m}\mid X=x)$ (i.e., $\hat{\gamma}_{m}^{y}(x)$) may not, on average, reflect the true underlying probabilities--that is, they may be miscalibrated. 
This can affect the performance of fair classifiers. 
As noted in Remark~\ref{rmk:sparsity}, our algorithms are estimator-agnostic, so existing calibration remedies can be directly plugged in when estimating group membership probabilities.
Thus, calibrated predictors can be used within our algorithms whenever needed.

To examine this, we use our DP postprocessing method as a representative example and compare it with a calibrated variant based on binning calibration. 
Specifically, we apply a binning-based calibrator, a standard approach for reducing miscalibration in probabilistic predictors, instantiated via Python's \texttt{functools.partial} with 40 bins and a prior strength of 1. Figure~\ref{apxfig:calibration} shows that the performance of the calibrated and uncalibrated versions is very close.

\begin{figure}[htbp]
 
\begin{center} 
    \begin{subfigure}{\textwidth} 
        \centering
        
        \begin{subfigure}{0.45\textwidth}
            \centering
            \includegraphics[width=\linewidth]{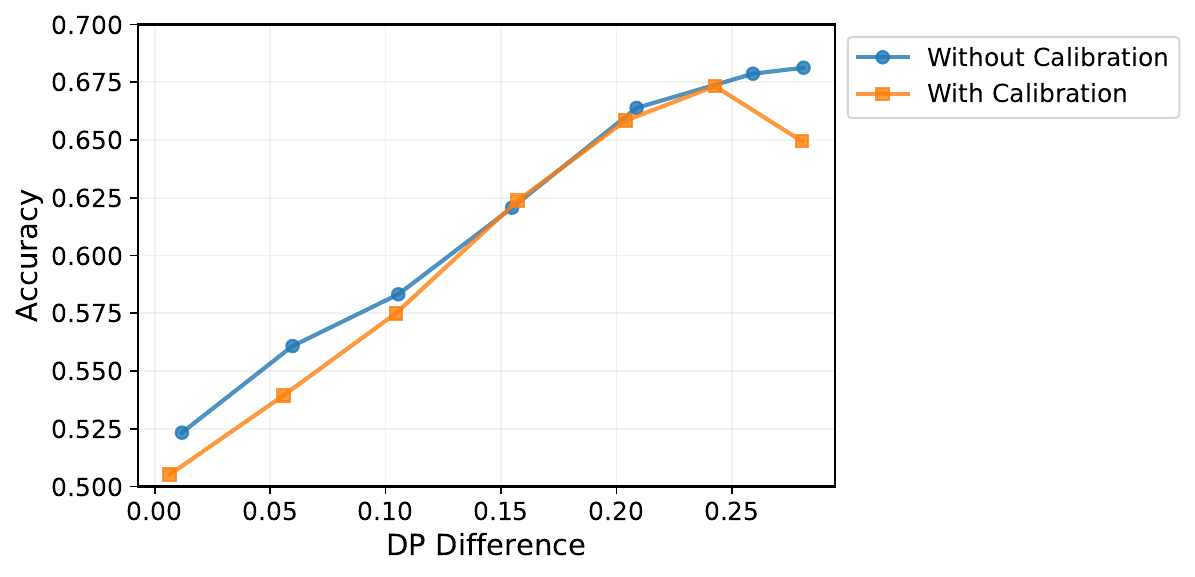}
            \caption{Our Postprocessing}
        \end{subfigure}
        \begin{subfigure}{0.45\textwidth}
            \centering
            \includegraphics[width=\linewidth]{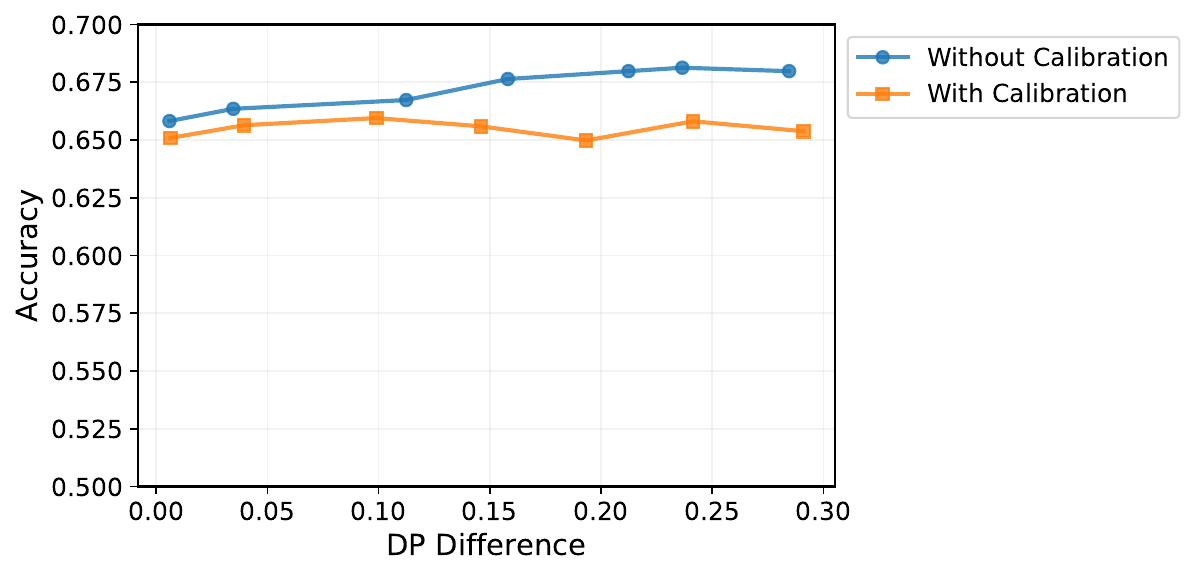}
            \caption{Our Postprocessing (aware)}
        \end{subfigure}

    \end{subfigure}
    \caption{Comparison between the uncalibrated and calibrated versions of our method on \texttt{COMPAS} when enforcing DP via MD.}
    \label{apxfig:calibration}
    \end{center}
    \vskip -0.1in
\end{figure}

}

\section{Selection of $\lambda$} \label{apxsec:lambda_selection}

In the experiment, we use grid searching as an example, following \citet{menon2018} and \citet{chen2024posthoc}.
For each $\lambda\in\Lambda$, we compute MD/MR and Accuracy on the validation set (size $N_v$), then pick highest-accuracy among ones with  MD/MR not greater than $\delta$ (complexity: $O(M|\Lambda|Nv)$). 
As noted in Remark \ref{rmk:lambda}, other tuning ways can replace grid search for $\lambda$ selection.
For example, dual update can be applied for more nuanced selection. 
Specifically, using projected subgradient ascent approach:
\begin{itemize}
    \item The fairness-constrained optimization problem is first transformed into its dual form, where the Lagrange multiplier $\lambda$ represents how strongly the fairness constraint influences the objective.
    \item In this dual formulation, one seeks  to maximize the dual objective with respect to $\lambda$ while minimizing the classifier's loss under the current $\lambda$.
    \item At each iteration, the classifier is updated or retrained based on the current $\lambda$, the fairness violation is measured on validation set, and $\lambda$ is adjusted in the direction of the gradient (or subgradient) that increases the dual objective--i.e., reduces the violation.
    \item Then, $\lambda$ is projected back into a bounded region to maintain stability, and this projected subgradient ascent continues until both fairness and accuracy reach equilibrium.
    The value of final $\lambda$ is returned. 
\end{itemize}


\section{Limitations and Future Directions}\label{apxsec:limitations}

{This paper contributes to the theoretical understanding of fair classification and supports the development of algorithms that reduce bias and enhance fairness in ML systems. 
Nonetheless, several limitations and future directions remain.

From the theoretical perspective, as with prior work \citep{menon2018,Zeng2022,zeng2024bayesoptimalfair}, our characterization of Bayes-optimal fair classifiers provides the best solution at the population level. 
Investigating the finite-sample properties and establishing corresponding guarantees remains an important future direction.
Additionally, while we follow \citet{menon2018,chen2024posthoc} in using grid search as an example implementation for selecting the value of $\lambda$, this approach can be computationally costly. 
As discussed in Remark \ref{rmk:lambda}, other efficient alternatives can be explored to improve solver efficiency.
Finally, both attribute-aware and attribute-blind settings in fair ML (so does our work) suppose the availability of sensitive feature during training.
However, in real-world scenarios, collecting and using sensitive information--even solely during the training phase--may raise privacy concerns and thus be restricted. 
This motivates future research on methods that handle partially available or entirely missing sensitive features during training \citep{Kallus2022}, and on integrating such approaches into our framework.
}}